\documentclass[10pt,twocolumn]{article}

\usepackage[margin=2cm]{geometry}
\usepackage{amsmath,amsfonts,amssymb}
\usepackage{newtxtext,newtxmath}
\usepackage{graphicx}
\usepackage{booktabs}
\usepackage{multirow}
\usepackage{makecell}
\usepackage{array}
\usepackage{tabularx}
\usepackage[numbers,super,sort&compress]{natbib}

\usepackage{url}
\usepackage[colorlinks=true,linkcolor=blue,citecolor=blue,urlcolor=blue]{hyperref}
\graphicspath{{img/}{./img/}}

\title{\bfseries S3VD: Semantic-Guidance Spatio-Temporal Scanning for \\ Video Deraining}

\author{Kui Jiang$^{1,2}$
\and Yi'ang Chen$^{1}$
\and Yan Luo$^{1}$
\and Zhaocheng Yu$^{1}$
\and Junjun Jiang$^{1,*}$
\and Xianming Liu$^{1}$}

\date{}

\begin{document}

\maketitle

\begin{abstract}
Heavy rainfall severely degrades outdoor videos by corrupting high-frequency details and introducing motion blur, critically undermining the reliability of visual tasks. Recently, State Space Models (SSMs), particularly Mamba, have emerged as efficient alternatives for vision tasks with their linear complexity and ability to model long-range dependencies. However, when confronted with the poor visual representations in rainy videos, Mamba still faces difficulties in preserving the integrity of 2D spatial semantics and modeling 3D spatio-temporal correlations. To break these limitations, we introduce S3VD, a Semantic-Guidance Spatio-Temporal Scanning framework for video deraining, featuring two key innovations: Multi-Scale Semantic Fusion (MSSF) Module and Spatio-Temporal Scanning Fusion (STSF) Module. The former integrates temporal semantic priors from DINOv2 to guide precise feature representation and counteract the loss of local semantic context inherent to Mamba's 1D flatten operation, enhancing robustness against extreme degradation. The latter introduces a spatio-temporal scanning mechanism and devises a Decoupled-Gating Mamba (DG-Mamba) layer, which employs two independent gates to adaptively control preceding and subsequent contextual information within the input clip, optimizing intra-frame and inter-frame correlation modeling. Experiments on video deraining benchmarks demonstrate the superiority of S3VD, achieving state-of-the-art performance with an average 0.84 dB PSNR improvement over Mamba-based baselines.
\end{abstract}

\noindent\textbf{Keywords:} Video Deraining, Spatio-Temporal Scanning, Semantic Guidance

\renewcommand\thefootnote{}
\footnotetext{$^{1}$School of Computer Science and Technology, Harbin Institute of Technology, Harbin 150001, China. $^{2}$Zhengzhou Advanced Research Institute, Harbin Institute of Technology, Zhengzhou 450001, China. $^{*}$Corresponding author: Junjun Jiang (jiangjunjun@hit.edu.cn)}

%
\section{Introduction}
\label{sec:introduction}


Outdoor visual perception systems, including surveillance cameras, unmanned aerial vehicle navigation platforms, and autonomous driving sensors, face significant technical challenges due to video degradation in adverse weather. 
Rain streak accumulation and transient raindrop occlusion distort scene textures and disrupt motion continuity, propagating errors into downstream tasks, such as semantic segmentation~\cite{zhao2023lif, zhong2022rainy} and object detection~\cite{zhang2025frequency, dai2024object}. 
This creates an urgent need for robust video deraining algorithms to enhance environmental adaptability.

\begin{figure}[htbp]
\centering
\includegraphics[width=0.48\textwidth, keepaspectratio]{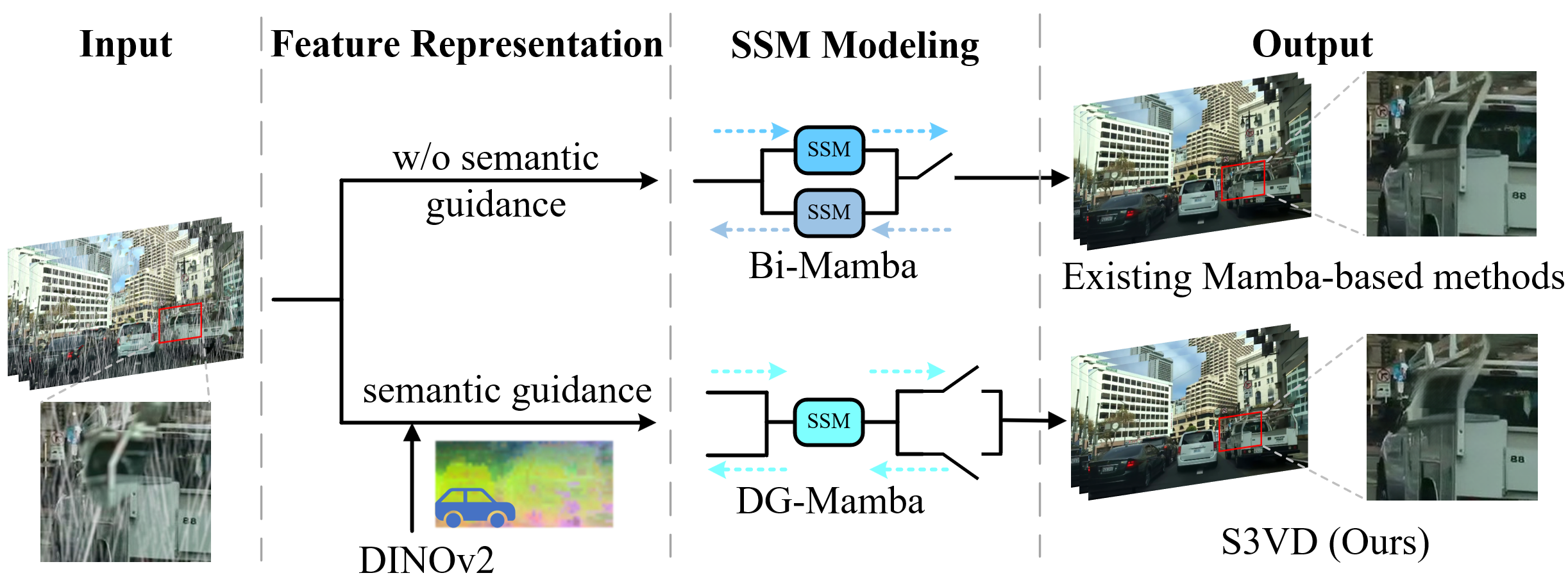}
\caption{Motivation for 
S3VD. 
Without semantic guidance, existing Mamba architecture (independent SSM streams + rigid fusion) struggles with complex scenes, causing texture loss (\emph{e.g.}, white truck). S3VD integrates semantic priors and a DG-Mamba layer with decoupled gates for adaptive fusion, effectively removing rain while preserving background integrity. 
}
\label{motivation}
\end{figure}

Early video deraining approaches~\cite{kim2015video, barnum2010analysis, ren2017video} relied on handcrafted priors (\emph{e.g.}, rain streak directionality, sparsity, or low-rank property) to separate rain artifacts from clean backgrounds. However, their manual parameter tuning and poor generalization hindered their performance in dynamically complex scenarios. 
While deep learning-based techniques have achieved remarkable success in image deraining~\cite{jiang2021rain, shi2026scale, jiang2025ph}, their principles reveal critical drawbacks in handling intricate spatio-temporal correlations in sequences~\cite{wu2023mask, liu2018d3r, guo2026align}. For example, while Convolutional Neural Networks (CNNs)~\cite{chen2018robust, liu2018d3r} based methods excel at capturing intra-frame spatial details and develop 3D convolution and temporal separable convolution to explore inter-frame correlation, they are still constrained by a limited spatio-temporal receptive field. 
Recurrent Neural Networks (RNNs)~\cite{yang2019frame, xue2025asf} improve global temporal awareness, but their reliance on external alignment modules (\emph{e.g.}, optical flow estimation, and deformable convolutions) results in rigid spatio-temporal fusion, causing inevitable error propagation like temporal artifacts. 
While Transformers~\cite{yang2023video, liang2024vrt} achieve flexible spatio-temporal modeling through the attention mechanism, their quadratic complexity incurs a significant computational bottleneck.

State Space Models (SSMs)~\cite{kalman1960new} offer promising alternatives for vision tasks~\cite{liu2024vmamba,jiang2026vdmamba}. The Mamba architecture~\cite{gu2023mamba} 
captures long-range dependencies with linear complexity, achieving notable performance in computer vision. Several works have already explored applying this paradigm to video deraining. RainMamba~\cite{wu2024rainmamba} leverages Hilbert scanning to explore sequence-based and patch-wise information for video deraining, while VDMamba~\cite{sun2025semi} introduces a dual-branch network with a dynamic stacking filter for real-world deraining. 
Although considerable improvement has been achieved, these methods still face core limitations for video deraining. 
Its 1D sequential processing, which flattens a temporal video into a 1D sequence of pixels, compromises the holistic spatial semantics of the video and disrupts spatio-temporal correlations, complicating the restoration of fine-grained details.
Furthermore, while the existing bidirectional extension Bi-Mamba~\cite{zhu2024vision} tries to improve holistic spatial modeling, its single gating mechanism to control both forward and backward information simultaneously prevents adaptive fusion and aggravates the spatio-temporal redundancy in videos. 
These limitations cause suboptimal deraining output with visible artifacts and temporal distortions, as shown in Fig.~\ref{motivation}.

Recent vision foundation models such as CLIP~\cite{radford2021learning} and DINO variants~\cite{caron2021emerging, oquab2023dinov2} have revolutionized feature learning through large-scale pre-training. In particular, for self-supervised models, DINO and its variants excel at extracting degradation-resistant semantic features without labels. Due to powerful feature representations that are inherently robust to photometric and geometric distortions, foundation model priors~\cite{radford2021learning, oquab2023dinov2} have also been applied to image restoration tasks~\cite{luo2023controlling, lin2023multi, liu2025unified}, yielding substantial performance gains. This inspires us to ask the following question:

\textbf{Can the semantic and contextual priors of vision foundation models facilitate robust feature representation for precise video deraining and content recovery?}

\begin{figure}[!t]
\centering
\includegraphics[width=0.48\textwidth, keepaspectratio]{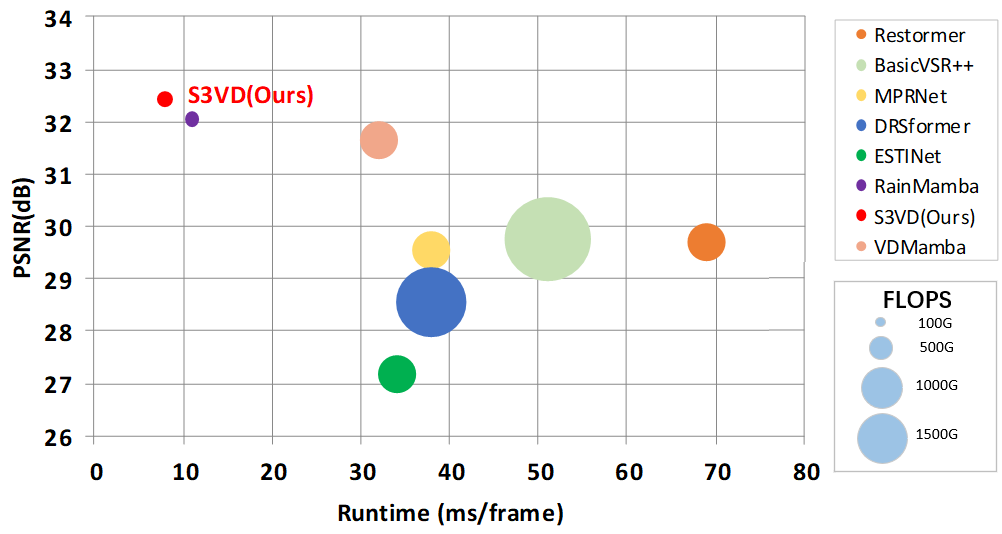}
\caption{FLOPs, Runtime and deraining performance comparison between our
S3VD and other state-of-the-art algorithms. 
S3VD achieves the highest PSNR with the fastest runtime and highly competitive computational efficiency among state-of-the-art methods. 
}
\label{cmp}
\end{figure}

Building upon these analyses above, we propose S3VD, a spatio-temporal Mamba framework enhanced with global semantic guidance. 
Specifically, we elaborate a Multi-Scale Semantic Fusion (MSSF) Module to integrate semantic priors from DINOv2~\cite{oquab2023dinov2} to promote feature representation. As a foundation model trained via self-supervision on massive-scale unlabeled data, DINOv2 yields feature priors with exceptional robustness and generalization against extreme degradations. 
By incorporating such semantic priors, MSSF provides high-quality semantic cues for content recovery. 
Building on this, we then tackle the spatio-temporal modeling by introducing a Spatio-Temporal Scanning Fusion (STSF) Module, which features spatial/temporal scanning mechanisms and a Decoupled-Gating Mamba (DG-Mamba) layer. DG-Mamba shares convolutional and SSM parameters while employing independent gates to adaptively fuse preceding and subsequent contextual information within the input clip, optimizing intra-frame (spatial) and inter-frame (temporal) correlation modeling.
Collectively, these enable efficient, high-fidelity rain distribution learning and background restoration. As illustrated in Fig.~\ref{cmp}, our S3VD achieves the highest PSNR with the fastest runtime while maintaining highly competitive computational efficiency compared to other state-of-the-art algorithms.

Overall, the main contributions of our proposed S3VD are summarized as follows.
\begin{itemize}
    \item We introduce a novel semantic-guidance spatio-temporal scanning architecture, called S3VD for video deraining, which effectively explores complex spatio-temporal correlations by integrating robust semantic priors. 
    \item We design a Multi-Scale Semantic Fusion (MSSF) Module to incorporate semantic priors from DINOv2, achieving a precise and robust feature representation. We also propose a Spatio-Temporal Scanning Fusion (STSF) Module that learns inter- and intra-frame correlations via distinct scanning mechanisms, featuring a novel Decoupled-Gating Mamba (DG-Mamba) layer for adaptive and effective spatio-temporal fusion.
    \item 
    Extensive experimental results on mainstream video deraining datasets demonstrate that our proposed S3VD achieves state-of-the-art performance, particularly excelling in temporal coherence and structural detail preservation across diverse rain scenarios.

\end{itemize}

\section{RELATED WORK}
\label{sec:relat}

In this section, we first review the recent advances in video deraining and pre-trained vision models. Then, we introduce the progress of state space models in vision tasks.

\subsection{Video Deraining}
Recent advances in deep learning have significantly propelled the field of video deraining. 
Early methods centered on using Convolutional Neural Networks (CNNs) to model the statistical properties of rain phenomena. MSCSC~\cite{li2018video} proposed a multiscale convolutional sparse coding  model to characterize the repetitive, multiscale nature of rain streaks. FastDeRain~\cite{jiang2018fastderain} utilized directional gradient priors to distinguish rain from the background. 
However, the local nature of convolutional limits the exploration of global context in video deraining.  
Researchers then began to seek more efficient alternatives with spatio-temporal modeling capability, like Recurrent Neural Networks (RNNs) and Transformers. FCRVD~\cite{yang2019frame} used a two-stage recurrent network with dual-level flow regularization to handle different rain. ESTINet~\cite{zhang2022enhanced} employed an LSTM network to aggregate spatio-temporal features across frames. VWR~\cite{wen2023video} suppressed artifacts and removed raindrops via spatio-temporal attention. ViMP-Net~\cite{wu2023mask} integrated mask-guided optical flow into dual transformer blocks to learn intra-frame spatial dependencies and inter-frame temporal context. While these methods brought improved performance, their heavy computational cost---stemming from the non-parallel processing of RNNs or the quadratic complexity of Transformers---remains a significant bottleneck. 
In contrast, our framework replaces recurrent architectures with a Mamba-based model, introducing bidirectional state propagation and adaptive gating to enhance long-range dependency modeling while retaining linear complexity.


\subsection{Pre-trained Vision Models}
The paradigm of vision foundation models like CLIP~\cite{radford2021learning} and DINO variants~\cite{caron2021emerging, oquab2023dinov2} has revolutionized feature learning through large-scale pre-training. By learning joint image-text embeddings through contrastive learning, CLIP has demonstrated remarkable zero-shot transfer capabilities for high-level tasks. For low-level vision, self-supervised models like the DINO variants are particularly relevant, as they excel at extracting degradation-resistant features without labels. By learning to match the features of a teacher and a student network on augmented views of the same image through self-distillation, DINO~\cite{caron2021emerging} develops powerful feature representations that are inherently robust to photometric and geometric distortions. DINOv2~\cite{oquab2023dinov2} further scales up this paradigm with improved loss functions and a significantly larger, curated dataset, resulting in enhanced robustness and generalization. 
These models enable cross-task adaptability, evidenced by DINO-IR’s~\cite{lin2023multi} success in multi-task image restoration. Building on these insights, we propose to integrate DINOv2 into video deraining, leveraging its global and robust semantic representation capability for degraded scenarios.

\subsection{State Space Models}
State space models (SSMs) like Mamba originated in Natural Language Processing (NLP) as an efficient architecture for 1D sequence modeling.  
Their ability to efficiently model long-range dependencies with linear complexity rapidly garnered significant interest in the vision domain. Pioneering works like VMamba~\cite{liu2024vmamba} and VisionMamba~\cite{zhu2024vision} successfully adapted Mamba to vision tasks by introducing multi-directional scanning and bidirectional architectures. And MambaIR~\cite{guo2024mambair} demonstrates its potential for low-level tasks. In the field of video deraining, some Mamba-based works like RainMamba~\cite{wu2024rainmamba} and VDMamba~\cite{sun2025semi} have gained improved performance. While these works rely on complex scanning strategies (\emph{e.g.}, RainMamba’s Hilbert scanning~\cite{wu2024rainmamba}) and elaborate external modules(\emph{e.g.}, VDMamba’s dynamic stacking filter~\cite{sun2025semi}), Mamba's inherent architectural limitations still remain unresolved: flattening 3D videos for 1D sequential processing causes significant loss of local semantic context and disrupts the inherent spatio-temporal structure. To tackle these issues, we propose a novel semantic-guidance spatio-temporal scanning network for video deraining, uniquely integrating robust semantic priors to guide the deraining process while redesigning Mamba's core architecture for efficient spatio-temporal modeling.

\section{Method} 
\label{sec:Method}
This section first introduces the overall architecture of the proposed S3VD and model optimization, followed by a detailed exposition of our Multi-Scale Semantic Fusion (MSSF) Module process and the implementation specifics of the Spatio-Temporal Scanning Fusion (STSF) Module.

\begin{figure*}[!ht]
\flushleft
\centering
\includegraphics[width=0.95\textwidth]{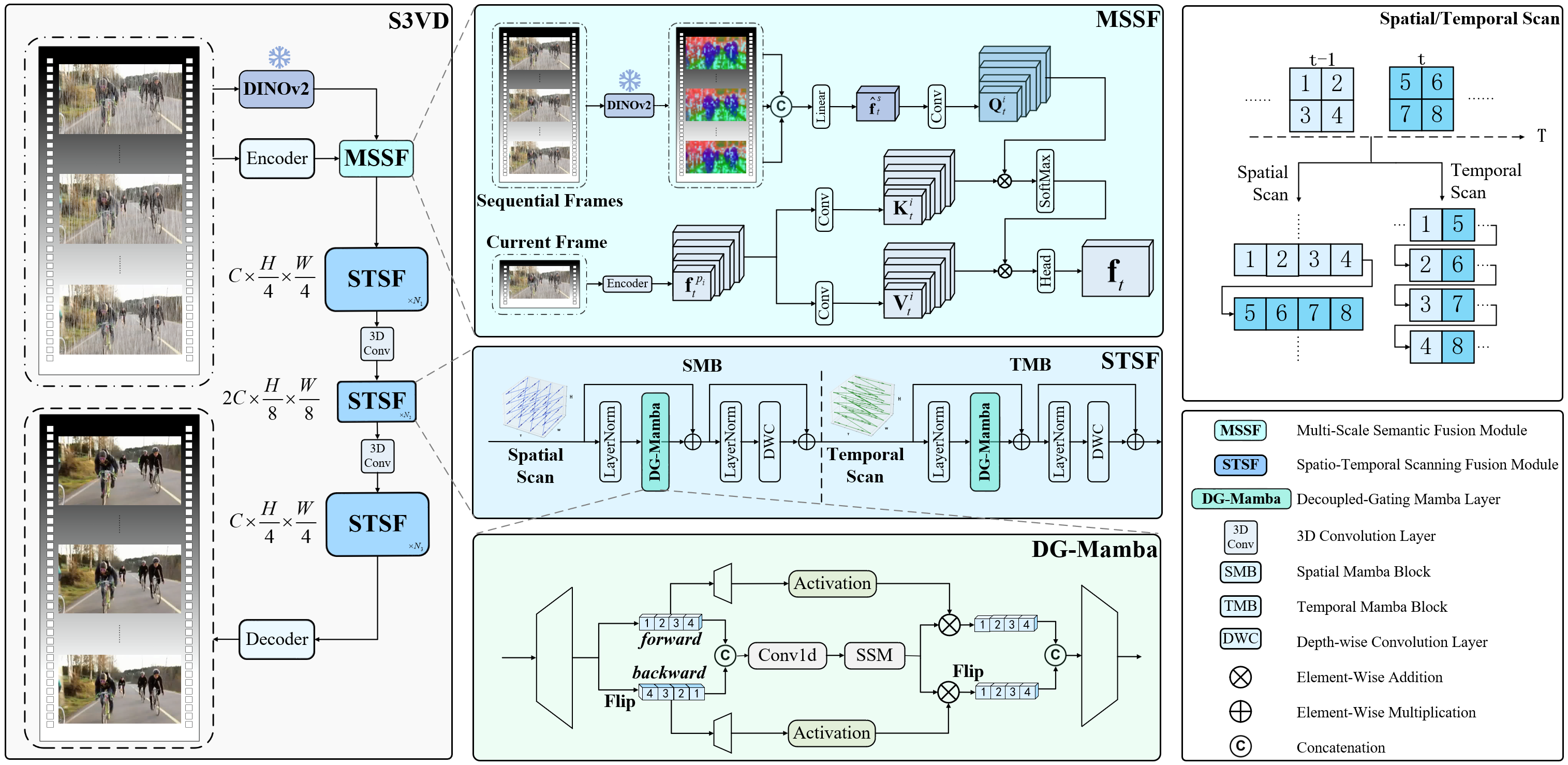}
\caption{Overall architecture of our S3VD framework for video deraining. The proposed network leverages DINOv2 with our Multi-Scale Semantic Fusion (MSSF) Module to fuse robust semantic representations and pixel features. Features are then processed through Spatio-Temporal Scanning Fusion (STSF) Modules, each consisting of a Spatial Mamba Block (SMB) and Temporal Mamba Block (TMB) for comprehensive spatio-temporal modeling. Each Mamba block incorporates our novel Decoupled-Gating Mamba (DG-Mamba) layer that enables efficient long-range dependency modeling in both spatial and temporal dimensions. 3D convolution layers bridge between STSF modules at different scales to ensure effective multi-scale feature integration before final reconstruction through the decoder.
}
\label{Framework}

\end{figure*}

\subsection{Architecture Overview and Optimization}
\label{sec:AMO}
\noindent \textbf{Architecture Overview.} Fig.~\ref{Framework} illustrates the overall framework of our Semantic-Guidance Spatio-Temporal Scanning architecture. Given a rainy video sequence $\{I_t\}_{t=1}^{T}$, where $I_t\in\mathbb{R}^{3 \times H \times W}$, we first employ two parallel feature extraction pathways to provide robust features for subsequent spatio-temporal modeling. Specifically, we utilize ConvNeXt~\cite{liu2022convnet} as an encoder for detailed pixel feature extraction while leveraging a frozen DINOv2~\cite{oquab2023dinov2} to capture robust semantic features. These complementary features are then processed through our novel MSSF, generating fused features $\{f_t\}_{t=1}^{T}$, where $f_t \in \mathbb{R}^{C \times H/4 \times W/4}$. Then, we implement serially arranged STSF Modules, positioned between two 3D convolutional blocks for downsampling and upsampling operations, to effectively learn rain streaks and raindrops of varying sizes and shapes. Each STSF integrates a Spatial Mamba Block followed by a Temporal Mamba Block, both incorporating our specially designed DG-Mamba layer. This architecture jointly models intra-frame spatial correlations and inter-frame temporal dependencies by aggregating bidirectional contextual information within the input clip. Finally, a reconstruction decoder composed of multiple 3D convolutional layers and upsampling layers generates the restored video frames $\{I^*_t\}_{t=1}^{T}$, where $I^*_t\in\mathbb{R}^{3 \times H \times W}$.

\noindent \textbf{Model Optimization.} 
To effectively train our video deraining network, we employ a comprehensive loss function. Firstly, we choose the pixel-level Charbonnier loss~\cite{charbonnier1994two} to ensure consistency between the output and ground truth in pixel space. The constraint is formulated as
\begin{equation}
    {\mathcal{L}_{pixel} = \frac{1}{N} \sum_{i=1}^N \sqrt{ \left(I^*_{t,i} - I^{gt}_{t,i}\right)^2 + \epsilon^2 }},
    \label{eq1}
\end{equation}
where the penalty coefficient $\epsilon$ is set to $10^{-12}$, and $I^*_{t,i}$ and $I^{gt}_{t,i}$ denote the $i$-th pixels of the restored and ground-truth frames at time $t$, respectively.
Secondly, we use the perceptual loss~\cite{johnson2016perceptual} to promote semantic consistency through high-level features extracted by the specific layers of a pre-trained VGG network, depicted as
\begin{equation}
    \mathcal{L}_{perceptual} = \mathcal{L}_{MSE} \Bigl(VGG_{3,8,15}(I^*_t), VGG_{3,8,15}(I^{gt}_t) \Bigr),
    \label{eq2}
\end{equation}
Finally, to enhance the model's ability to capture the spatio-temporal correlations, we utilize the dynamic contrastive learning loss~\cite{wu2024rainmamba}:
\begin{equation}
{\mathcal{L}_{DCL} = \frac{1}{S} \sum_{s=1}^{S} \left( \sum_{r=1}^{2} \frac{\mathcal{L}_{L_1} \left( G_r(P_{j,s}), G_r(I^*_{t,s}) \right)}{\mathcal{L}_{L_1} \left( G_r(N_{w,s}), G_r(I^*_{t,s}) \right)} \right)},
\label{eq3}
\end{equation}
where $G_r(\cdot)$ denotes the feature extractor at the $r$-th selected low-level layer of the pre-trained VGG-19 model~\cite{simonyan2014very}. The anchor $I^*_{t,s}$ is the $s$-th patch from output frame $t$. The positive sample $P_{j,s}$ is the spatially corresponding patch from a neighboring clean frame, where $j\in\{t-1,t,t+1\}$, while the negative sample $N_{w,s}$ is a spatially distant patch from a degraded frame, where $w\in\{1,\ldots,T\}$.
By pulling positive samples closer to and pushing negative ones apart from the anchors, DCL encourages the network to learn the spatio-temporal self-similarity within clean video frames, thereby improving the restoration of fine details.  
The total loss function is:
\begin{equation}
\mathcal{L}=\mathcal{L}_{pixel}+\lambda_1 \mathcal{L}_{perceptual}+\lambda_2 \mathcal{L}_{DCL},
\label{eq4}
\end{equation}
where the weight hyper-parameters $\lambda_1$ and $\lambda_2$ are empirically set to 0.3 and 0.1.

\subsection{Multi-Scale Semantic Fusion Module}
While Mamba-based architectures are efficient for sequence modeling, they often sacrifice spatial semantic integrity, resulting in poor handling of complex spatial structures, especially when heavy rain frequently causes severe degradation in pixel-level representations.
To address this, we propose the Multi-Scale Semantic Fusion (MSSF) Module that incorporates robust semantic priors extracted from DINOv2 to guide pixel feature learning, as Fig.~\ref{Framework} shows. The MSSF operates through two complementary branches that work in concert to achieve comprehensive semantic-pixel feature integration. Since heavy rain can make local pixel cues unreliable, DINOv2 semantics guide the ConvNeXt pixel stream through cross-attention rather than replace it, while the residual connection in Eq.~\eqref{eq10} preserves the pixel features required for color and detail reconstruction.

\noindent{\textbf{Semantic Branch.}}
While DINOv2 excels at extracting rich, structure-aware semantic representations, its static nature, stemming from the image-based pre-training, presents a critical limitation when confronted with the complex spatio-temporal dynamics of rainy videos. The compounded effects of motion blur and rain interference can compromise the integrity of a single frame's feature extraction, thus limiting the completeness of the resulting semantic representation.

To mitigate this issue, we propose to enhance the semantic representation of each frame by incorporating temporal context from adjacent frames. The key idea is to allow semantic tokens to aggregate contextual cues across time, so that corrupted regions in the current frame can be supplemented by cleaner or more stable observations from neighboring frames.

For the input video $\{I_t\}_{t=1}^{T}$, we first use a frozen DINOv2 model to extract its per-frame semantic features:
\begin{equation}
\hat{f}^s_t = \mathcal{F}_{\mathrm{DINOv2}}(I_t).
\label{eq5}
\end{equation}
Then, to incorporate temporal awareness, we concatenate each frame's semantic feature alongside the channel dimension:
\begin{equation}
f^{s}_t = \mathcal{W}\left(\mathrm{Concat}\left\{\hat{f}^s_1, \ldots,\hat{f}^s_t, \ldots, \hat{f}^s_T\right\}\right),
\label{eq6}
\end{equation}
where $\mathcal{W}$ is a linear projection layer that maps the concatenated feature back to the original channel size, and $f^{s}_t$ denotes the final extracted strong semantic feature map.

\noindent{\textbf{Pixel Branch.}}
While semantic priors offer high-level guidance, spatial restoration relies heavily on low-level pixel features that capture fine-grained details. To this end, we employ a hierarchical ConvNeXt~\cite{liu2022convnet} encoder to extract pixel-level features at four different resolution levels:
\begin{equation}
\{f_t^{p_1}, f_t^{p_2}, f_t^{p_3}, f_t^{p_4}\} = \mathcal{F}_{\mathrm{ConvNeXt}}(I_t).
\label{eq7}
\end{equation}
These four levels correspond to progressively deeper layers in the encoder, and naturally represent features with increasing receptive fields and abstraction levels. Since rain affects video frames across multiple spatial scales---from subtle texture erosion to large-scale occlusion---capturing and refining information at all levels is essential.

\noindent{\textbf{Multi-Scale Semantic-Pixel Fusion.}}
To align high-level semantic understanding with low-level pixel restoration, we fuse temporally enriched semantic features with multi-scale pixel features via a cross-attention mechanism.

Injecting semantic priors into pixel representation is critical for two reasons. First, semantic features are more robust to visual degradation and can identify important object-level regions even under heavy rain. Second, pixel features from different layers of the encoder capture spatial details at different levels---from textures to shapes---which require complementary semantic context for accurate recovery.

To enable targeted guidance, we treat semantic features as queries and pixel features as keys and values in a scale-wise attention module. Specifically, we project the temporal semantic embedding $f_t^{s}$ to four resolutions $\{f_t^{s_1}, f_t^{s_2}, f_t^{s_3}, f_t^{s_4}\}$ via $\mathcal{F}_{\mathrm{Conv}}(\cdot)$.
Each $f_t^{s_i}$ interacts with pixel feature maps $f_t^{p_i}$ using the following cross-attention formulation:
\begin{equation}
Q_t^i = W_{Q_i} f_t^{s_i}, \quad
K_t^i = W_{K_i} f_t^{p_i}, \quad
V_t^i = W_{V_i} f_t^{p_i},
\label{eq9}
\end{equation}
\begin{equation}
f_t^i = \mathrm{Softmax}\left(\frac{Q_t^i K_t^{i\mathrm{T}}}{\sqrt{C}}\right) V_t^i + f_t^{p_i}.
\label{eq10}
\end{equation}
This mechanism allows the model to emphasize semantically relevant areas at each spatial scale, promoting structure-aware and rain-resilient feature learning.

Finally, we unify the four scale-aware outputs using a head with a projection layer and bilinear interpolation:
\begin{equation}
f_t = \mathcal{F}_{\mathrm{Head}}(f_t^1, f_t^2, f_t^3, f_t^4).
\label{eq11}
\end{equation}
The output $f_t$ serves as the encoder result passed to the subsequent STSF module, providing a semantically enriched, temporally guided representation.

\subsection{Spatio-Temporal Scanning Fusion Module}
Video deraining demands not only accurate spatial detail recovery but also temporal consistency across frames. Although Mamba offers efficient sequence modeling, its 1D flattening process lacks explicit handling of spatio-temporal heterogeneity in 3D video sequences.

\noindent\textbf{STSF.}
To address this limitation, STSF sequentially applies a Spatial Mamba Block (SMB) and a Temporal Mamba Block (TMB) that share the same architecture but use different traversal orders. For the stacked feature tensor $\mathbf{F}=[f_1,\ldots,f_T]\in\mathbb{R}^{T\times C\times H\times W}$, their zero-based traversal indices are defined as
\begin{equation}
{
\begin{aligned}
\pi_{\mathrm{S}}(t,h,w) &= tHW+hW+w,\\
\pi_{\mathrm{T}}(t,h,w) &= hWT+wT+t,
\end{aligned}}
\label{eq:traversal_orders}
\end{equation}
where $0\leq t<T$, $0\leq h<H$, and $0\leq w<W$. SMB scans spatial positions within each frame before advancing in time, whereas TMB scans time first at each spatial position. The backward branch reverses each order, and an inverse permutation restores the original tensor layout.
\begin{equation}
v_t= \mathcal{F}_\mathrm{DG-Mamba}(\mathcal{F}_\mathrm{LN}(f_{t}))+f_t,
\label{eq12} 
\end{equation}
\begin{equation}
\mathcal{F}_\mathrm{SMB/TMB}(v_{t})= \mathcal{F}_\mathrm{DWC}(\mathcal{F}_\mathrm{LN}(v_t))+v_t.
\label{eq13} 
\end{equation}

\noindent\textbf{DG-Mamba.}
Compared with Bi-Mamba~\cite{zhu2024vision}, which uses separately parameterized forward and backward Conv1d/SSM branches with a common gating formulation and element-wise fusion, DG-Mamba shares the Conv1d/SSM core, employs direction-specific gates, and concatenates the two gated responses before linear projection. This design reduces duplicated sequence modeling while retaining direction-dependent feature selection.

\begin{table*}[!htb]
\centering
\caption{Quantitative comparison with state-of-the-art methods on four benchmarks. The best results are highlighted in \textbf{bold}.}
\label{tab:cmp}
\renewcommand{\arraystretch}{1}
\setlength{\tabcolsep}{1.2pt}
\resizebox{1.00\linewidth}{!}{
\begin{tabular}{ll|ll|ll|ll}
\toprule
\multicolumn{2}{c|}{\textbf{VRDS}} &
\multicolumn{2}{c|}{\textbf{RainVID\&SS}} &
\multicolumn{2}{c|}{\textbf{RainSynAll100}} &
\multicolumn{2}{c}{\textbf{LWDDS}} \\
\cmidrule(lr){1-2} \cmidrule(lr){3-4} \cmidrule(lr){5-6} \cmidrule(lr){7-8}

Method & PSNR$\uparrow$/SSIM$\uparrow$/LPIPS$\downarrow$ &
Method & PSNR$\uparrow$/SSIM$\uparrow$ &
Method & PSNR$\uparrow$/SSIM$\uparrow$ &
Method & PSNR$\uparrow$/SSIM$\uparrow$ \\
\midrule

CCN~\cite{quan2021removing} & 23.75 / 0.8410 / 0.2091 &
FastDerain~\cite{jiang2018fastderain} & 17.88 / 0.4506 &
FCRVD~\cite{yang2019frame} & 21.06 / 0.7405 &
CCN~\cite{quan2021removing} & 27.53 / 0.9220 \\

ESTINet~\cite{zhang2022enhanced} & 27.17 / 0.8436 / 0.2253 &
MSCSC~\cite{li2018video} & 19.19 / 0.5250 &
RMFD~\cite{yang2021recurrent} & 25.14 / 0.9172 &
VWR~\cite{wen2023video} & 30.72 / 0.9726 \\

DRSformer~\cite{chen2023learning} & 28.54 / 0.9075 / 0.1143 &
PReNet~\cite{ren2019progressive} & 24.90 / 0.7464 &
BasicVSR++~\cite{chan2022basicvsr++} & 27.67 / 0.9135 &
BasicVSR++~\cite{chan2022basicvsr++} & 32.37 / 0.9792 \\

BasicVSR++~\cite{chan2022basicvsr++} & 29.75 / 0.9171 / 0.1023 &
S2VD~\cite{yue2021semi} & 29.69 / 0.9136 &
NCFL~\cite{huang2022neural} & 28.11 / 0.9235 &
ViMP-Net~\cite{wu2023mask} & 34.22 / 0.9784 \\

ViMP-Net~\cite{wu2023mask} & 31.02 / 0.9283 / 0.0862 &
VDMamba~\cite{sun2025semi} & 32.69 / 0.9233 &
SALN~\cite{ye2023sequential} & 29.78 / 0.9315 &
SALN~\cite{ye2023sequential} & 36.57 / 0.9802 \\

VDMamba~\cite{sun2025semi} & 31.65 / 0.9379 / 0.0733 &
MPEVNet~\cite{sun2023event} & 33.47 / 0.9391 &
VDMamba~\cite{sun2025semi} & 32.01 / 0.9428 &
DeLiVR~\cite{sun2025delivr} & 36.72 / 0.9801 \\

DeLiVR~\cite{sun2025delivr} & 31.98 / 0.9405 / 0.0624 &
DeLiVR~\cite{sun2025delivr} & 33.87 / 0.9420 &
RainMamba~\cite{wu2024rainmamba} & 32.16 / 0.9446 &
RainMamba~\cite{wu2024rainmamba} & 37.21 / 0.9816 \\

RainMamba~\cite{wu2024rainmamba} & 32.04 / 0.9366 / 0.0684 &
RainMamba~\cite{wu2024rainmamba} & 34.39 / 0.9496 &
DeLiVR~\cite{sun2025delivr} & 33.37 / 0.9419 &
VDMamba~\cite{sun2025semi} & 37.33 / 0.9821 \\

\midrule
\textbf{S3VD (Ours)} & \textbf{32.43 / 0.9427 / 0.0610} &
\textbf{S3VD (Ours)} & \textbf{35.19 / 0.9583} &
\textbf{S3VD (Ours)} & \textbf{34.02 / 0.9487} &
\textbf{S3VD (Ours)} & \textbf{38.07 / 0.9824} \\

\bottomrule
\end{tabular}
}
\end{table*}

Following the spatial or temporal rearrangement defined above, the stacked tensor $\mathbf{F}$ is flattened into a 1D sequence $x$ and processed by DG-Mamba as follows:
\begin{align}
    x_\mathrm{forward} & = \mathcal{F}_\mathrm{SSM}(\mathcal{F}_\mathrm{Conv1d}(\mathcal{W}_1(x))),\\
    x_\mathrm{backward} & = \mathcal{F}_\mathrm{SSM}(\mathcal{F}_\mathrm{Conv1d}(\mathrm{Flip}(\mathcal{W}_1(x)))),\\
    z_\mathrm{forward} & = \mathcal{F}_\mathrm{SiLU}(\mathcal{W}_2(\mathcal{W}_1(x))),\\
    z_\mathrm{backward} & = \mathcal{F}_\mathrm{SiLU}(\mathcal{W}_3(\mathrm{Flip}(\mathcal{W}_1(x)))),\\
    y & = \mathcal{W}_4( \mathrm{Concat}\{ (z_{\mathrm{forward}} \mathbin{\odot} x_{\mathrm{forward}}), \nonumber \\
      & \qquad \mathrm{Flip}(z_{\mathrm{backward}} \mathbin{\odot} x_{\mathrm{backward}}) \}),
\end{align}
where $\{\mathcal{W}_i\}_{i=1}^{4}$ denotes the four linear layers, $x_\mathrm{forward}$ and $x_\mathrm{backward}$ represent the forward and backward sequences, respectively, and $z_\mathrm{forward}$, $z_\mathrm{backward}$ serve as gate-controlled variables for each sequence. $y$ is the final processed 1D sequence. For the temporal scan, the forward branch aggregates context from preceding frames, whereas the backward branch explicitly incorporates information from subsequent frames through sequence reversal. Thus, DG-Mamba performs bidirectional, non-causal modeling within the available input clip rather than causal sequence modeling. We then reshape $y$ back to its original dimension for deep extraction. 

\section{Experimental Results}
\label{sec:EX}
To validate our proposed S3VD, we conducted experiments on four video deraining benchmark datasets, comparing our method against state-of-the-art approaches. We employed three evaluation metrics—Peak Signal-to-Noise Ratio (PSNR), Structural Similarity Index (SSIM), and Learned Perceptual Image Patch Similarity (LPIPS)—to quantitatively compare the performance between our method and various alternatives.

\subsection{Implementation Details}
\label{sec:ID}
\noindent \textbf{Datasets.}
Video Raindrops and Rain Streaks~\cite{wu2023mask} (\textbf{VRDS}) is the first synthetic video dataset that simultaneously features both raindrops and rain streaks. It encompasses various driving scenarios, including urban areas, villages, highways, and natural environments. The VRDS dataset consists of 102 videos in total, with each video containing 100 frames. The training set includes 72 videos, while the remaining 30 videos constitute the test set.

Rain Video Detection and Semantic Segmentation~\cite{sun2023event} (\textbf{RainVID\&SS}) is a synthetic video dataset with only rain streaks. It includes short clips from the ImageNet-NID dataset for video object detection and long clips from the Cam-Vid dataset for video semantic segmentation. The training set comprises 205 short clips and 3 long clips, while the test set includes 86 short clips and 2 long clips.

\textbf{RainSynAll100}~\cite{yang2021recurrent} is a synthetic video dataset that exclusively features rain streaks. Its training set consists of 900 videos, while the test set includes 100 videos.

Large-scale Waterdrop Dataset for Driving Scenes~\cite{wen2023video} (\textbf{LWDDS}) is the first synthetic video dataset focused exclusively on raindrops in driving scenarios. The training set consists of 45 videos with a total of 67,500 frames, while the test set contains 6 videos with 600 frames.  

\begin{figure*}[t]
    \centering
    \renewcommand\arraystretch{0.25}
    \setlength{\tabcolsep}{0pt}

    \textbf{(a) Visual comparison on the VRDS dataset~\cite{wu2023mask}.}\\[2pt]

    \begin{tabular}{@{}*{8}{c}@{}}
        \includegraphics[width=0.125\textwidth]{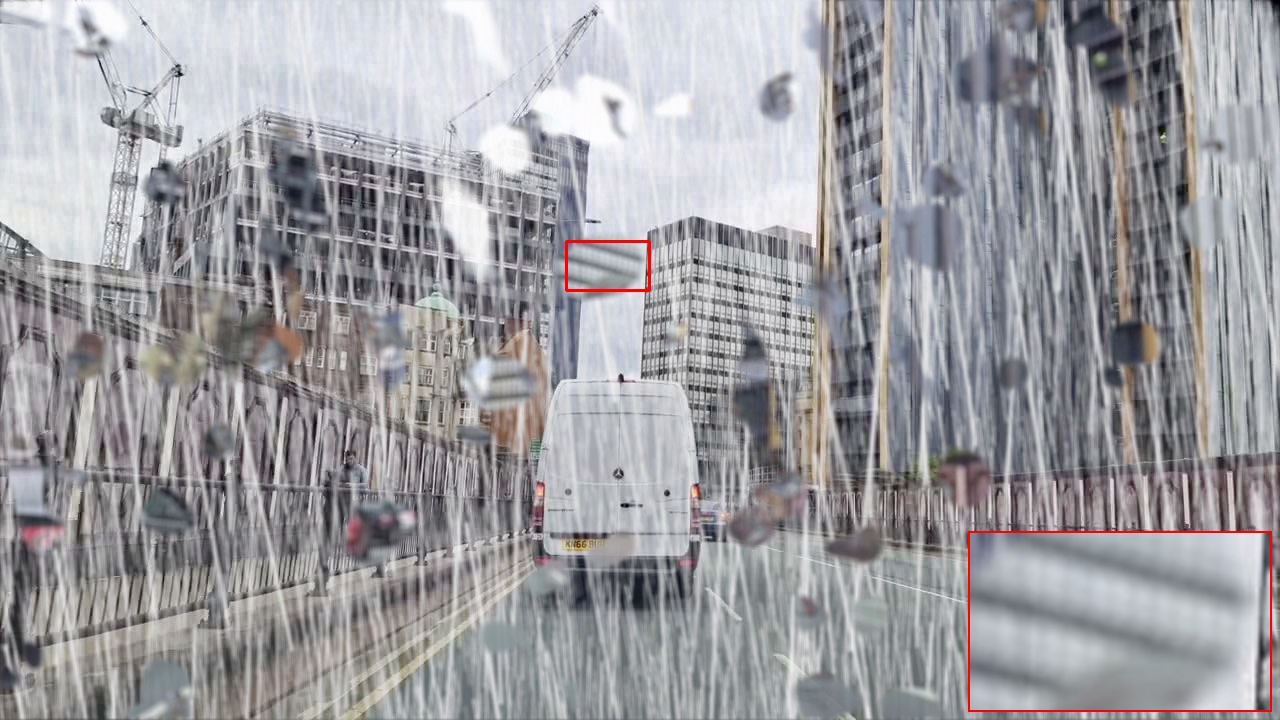} &
        \includegraphics[width=0.125\textwidth]{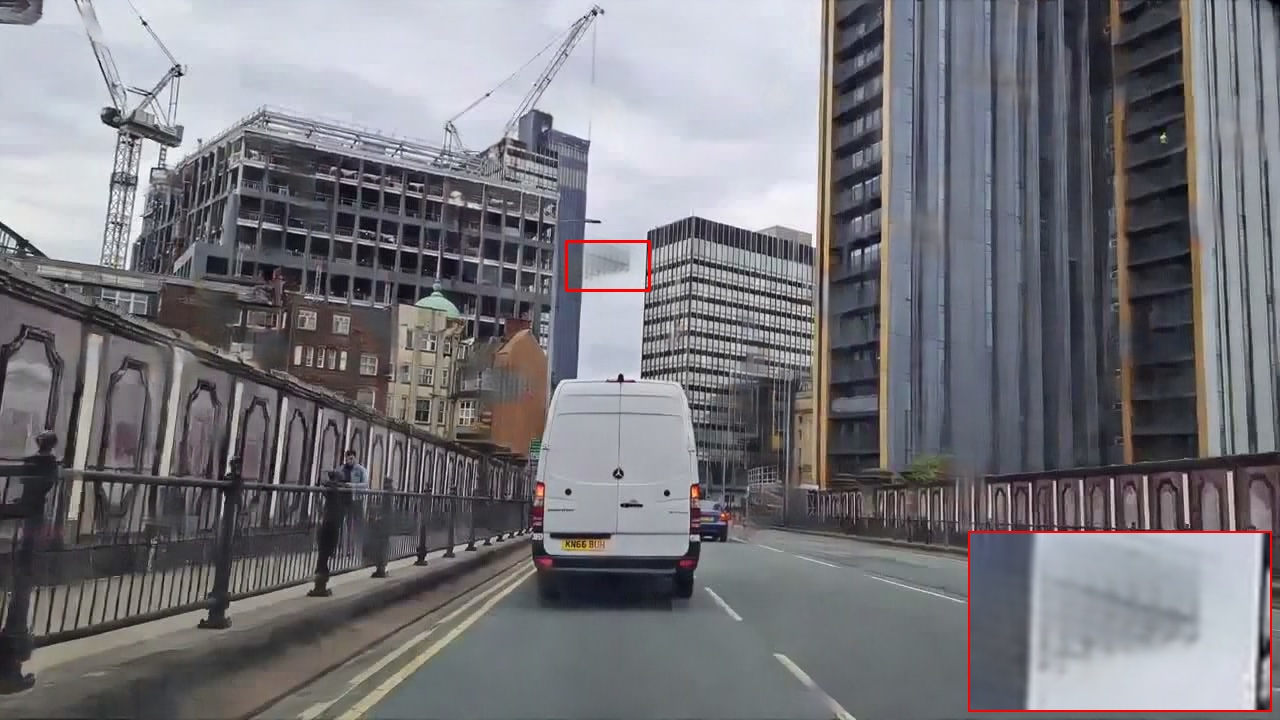} &
        \includegraphics[width=0.125\textwidth]{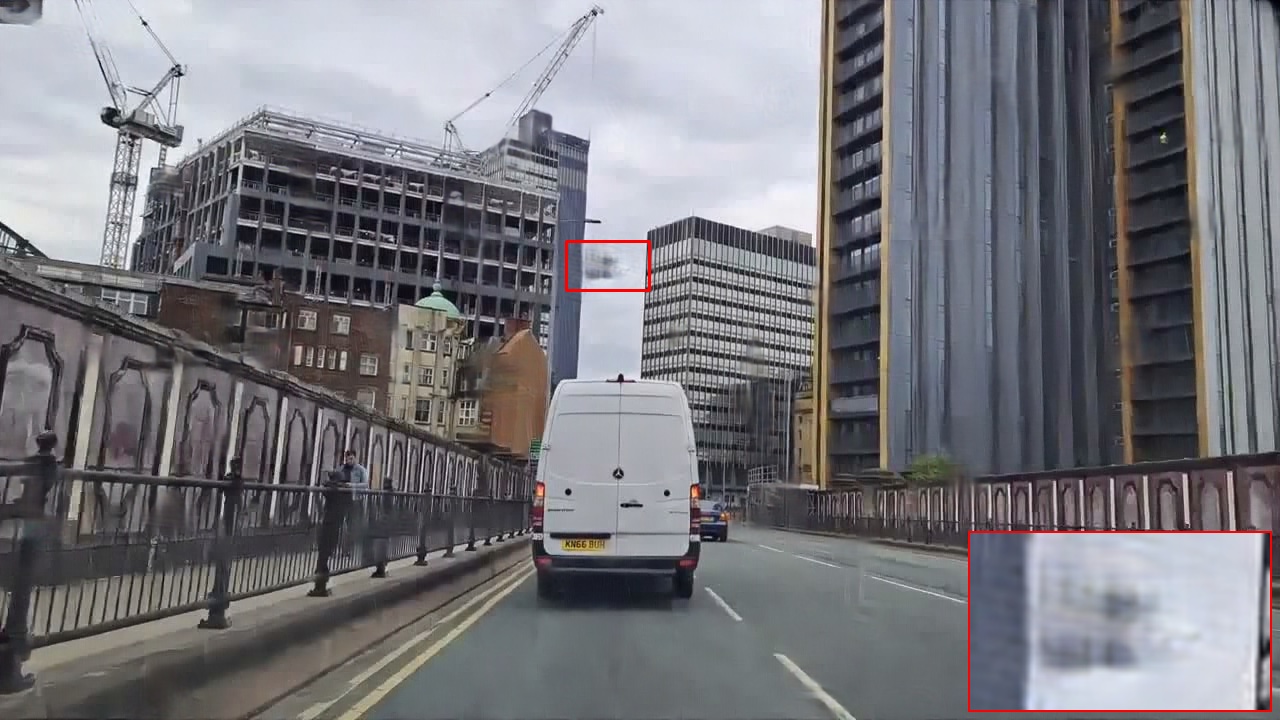} &
        \includegraphics[width=0.125\textwidth]{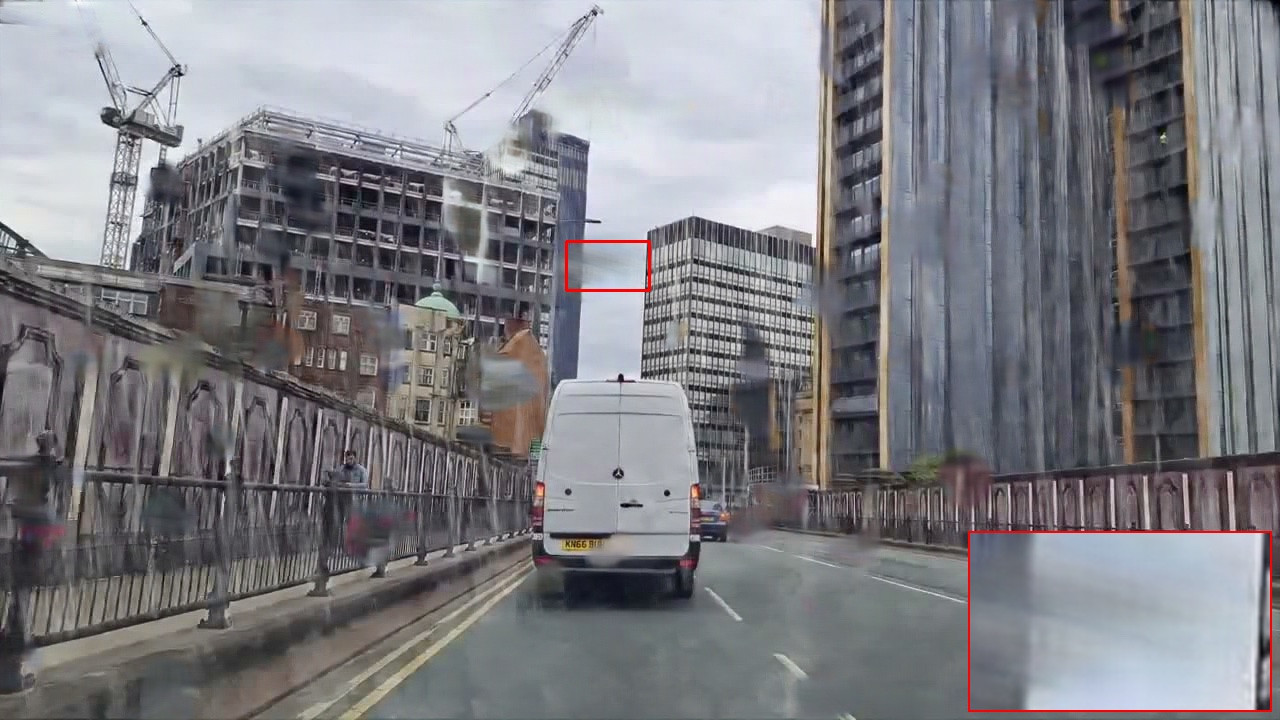} &
        \includegraphics[width=0.125\textwidth]{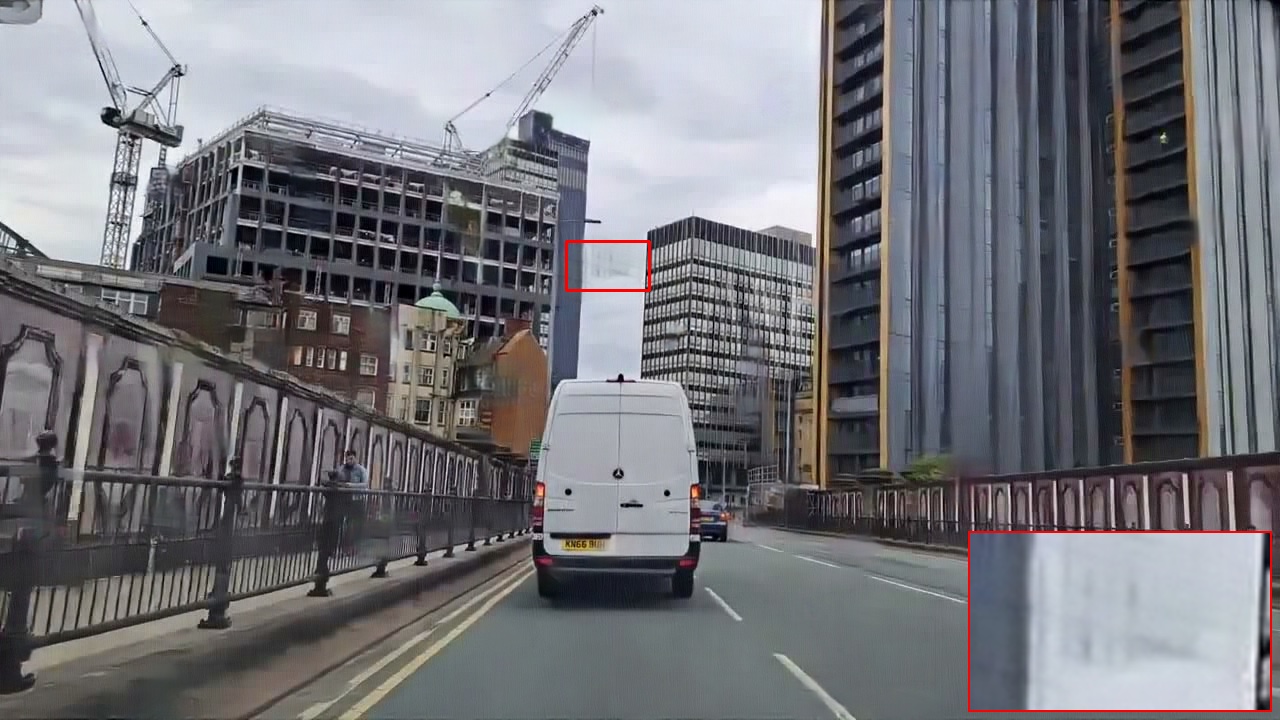} &
        \includegraphics[width=0.125\textwidth]{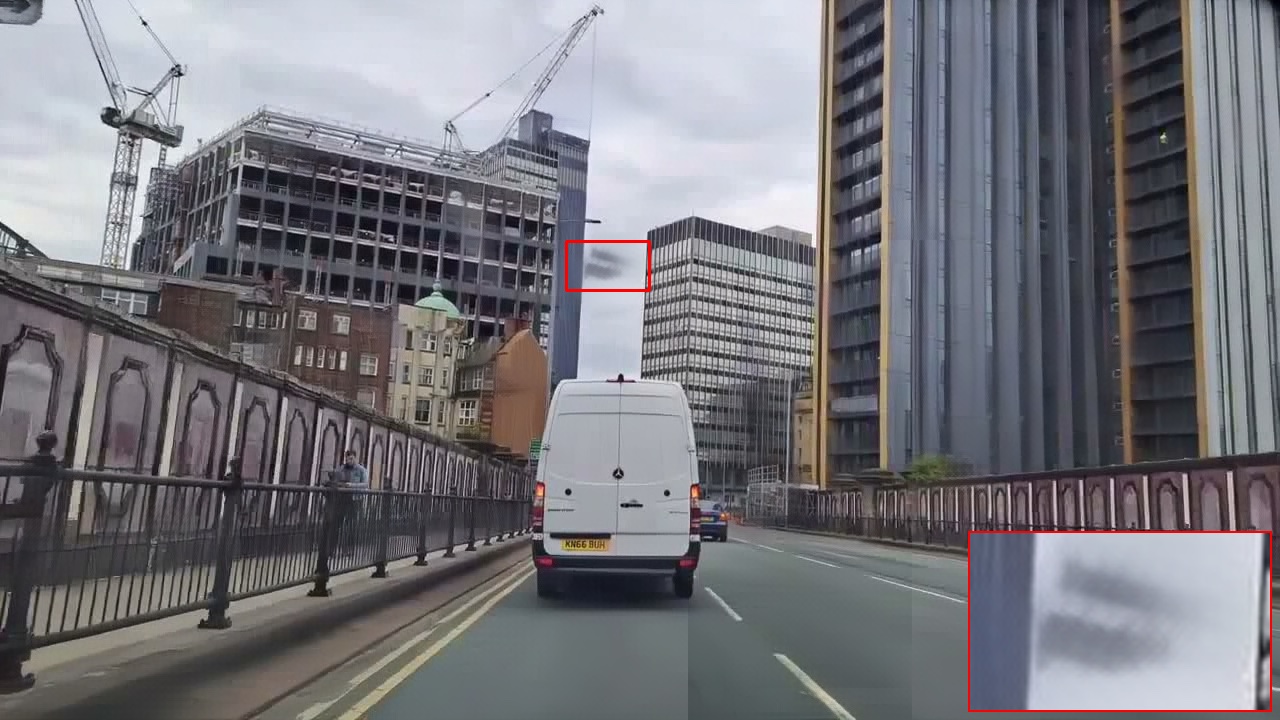} &
        \includegraphics[width=0.125\textwidth]{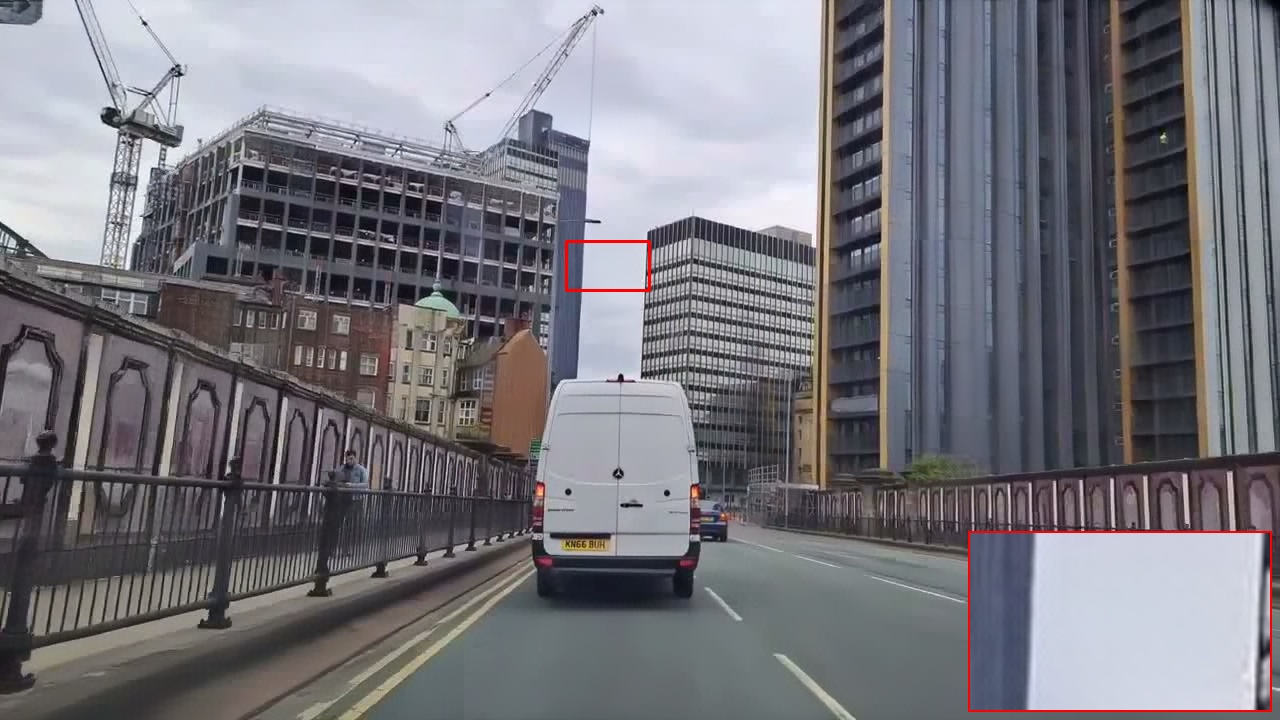} &
        \includegraphics[width=0.125\textwidth]{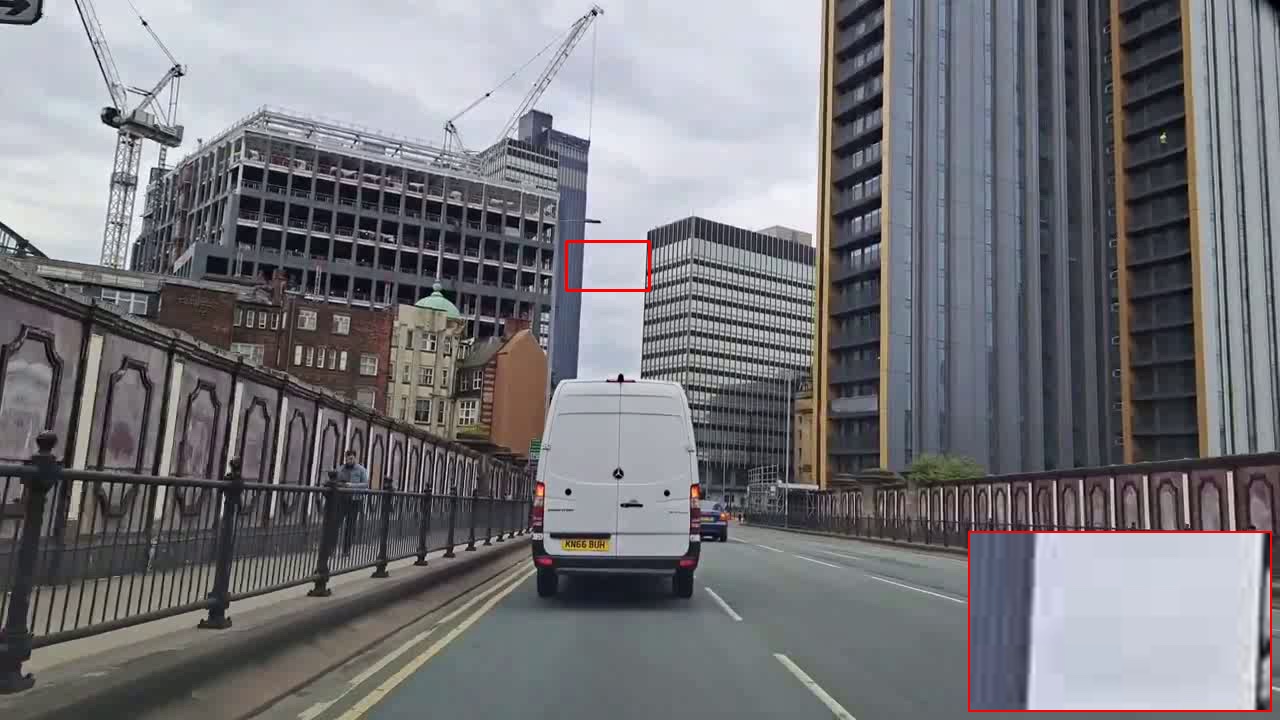} \\

        \includegraphics[width=0.125\textwidth]{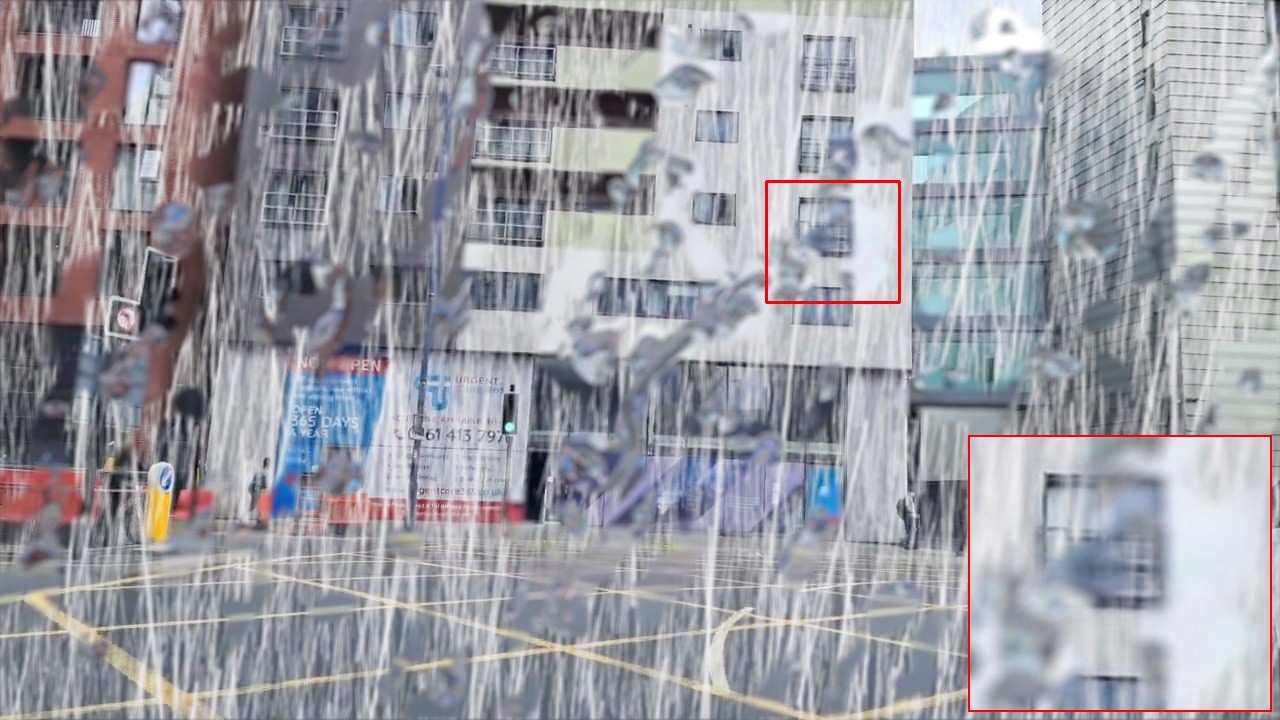} &
        \includegraphics[width=0.125\textwidth]{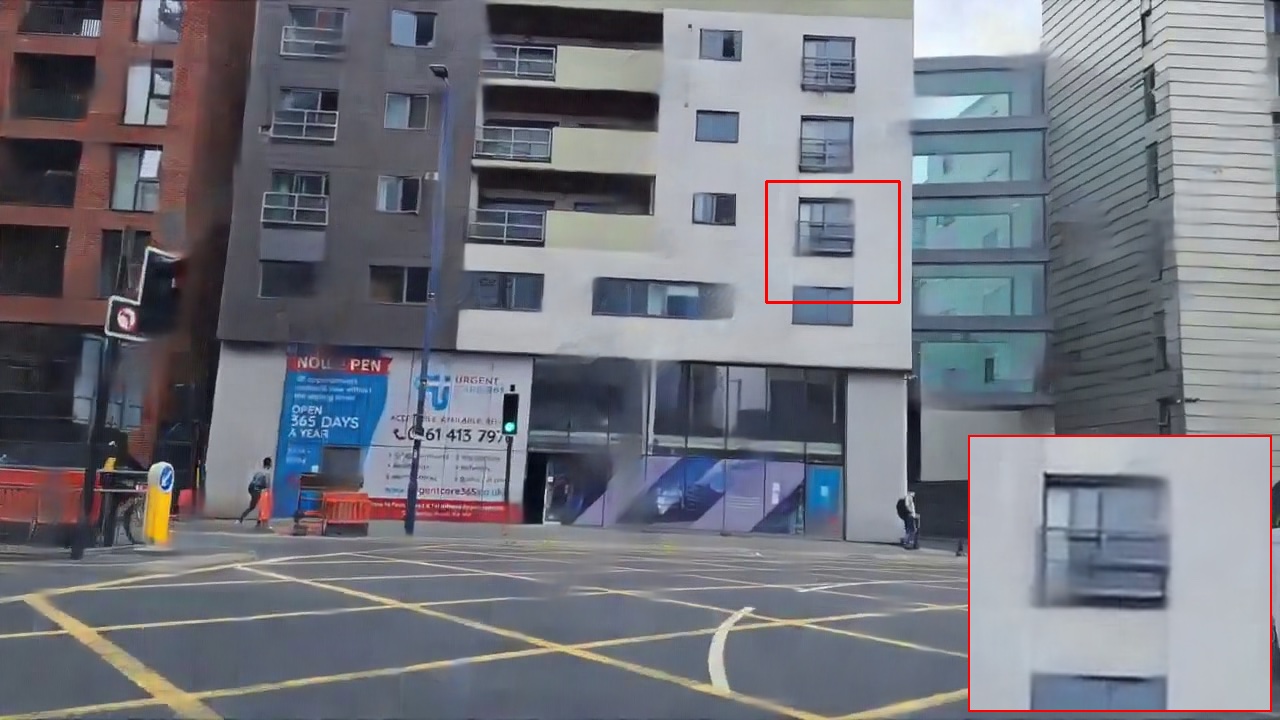} &
        \includegraphics[width=0.125\textwidth]{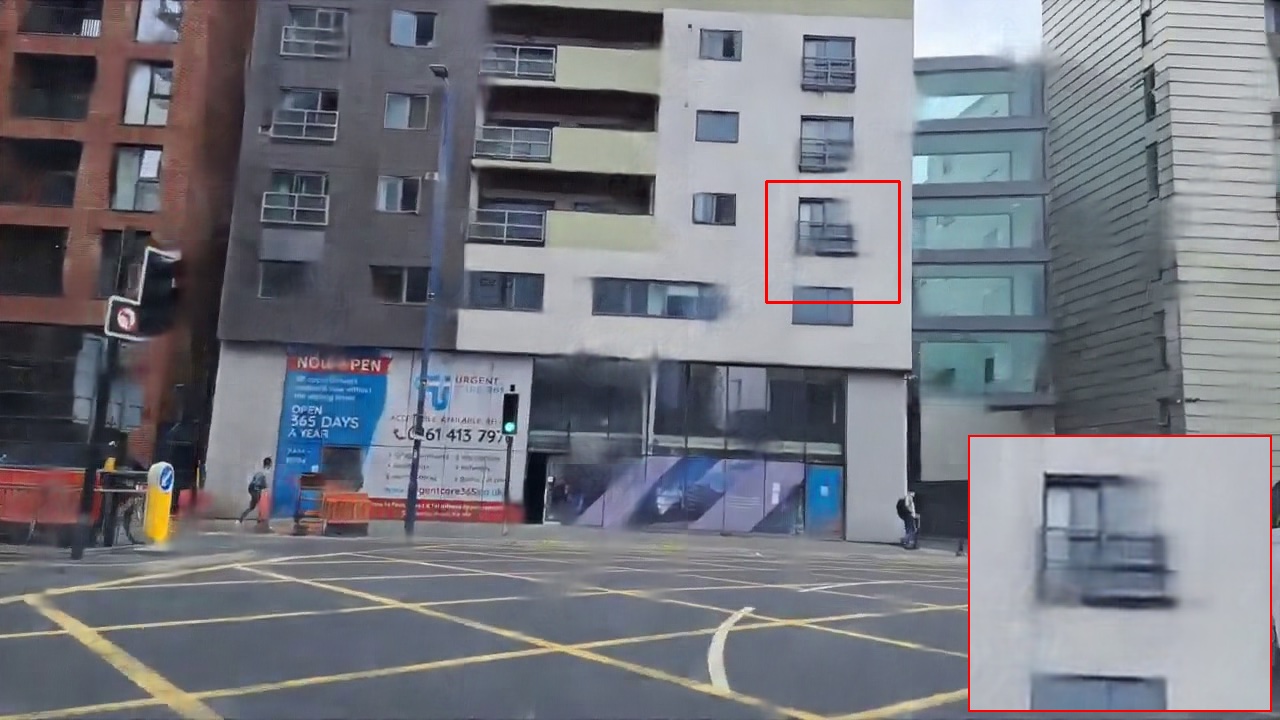} &
        \includegraphics[width=0.125\textwidth]{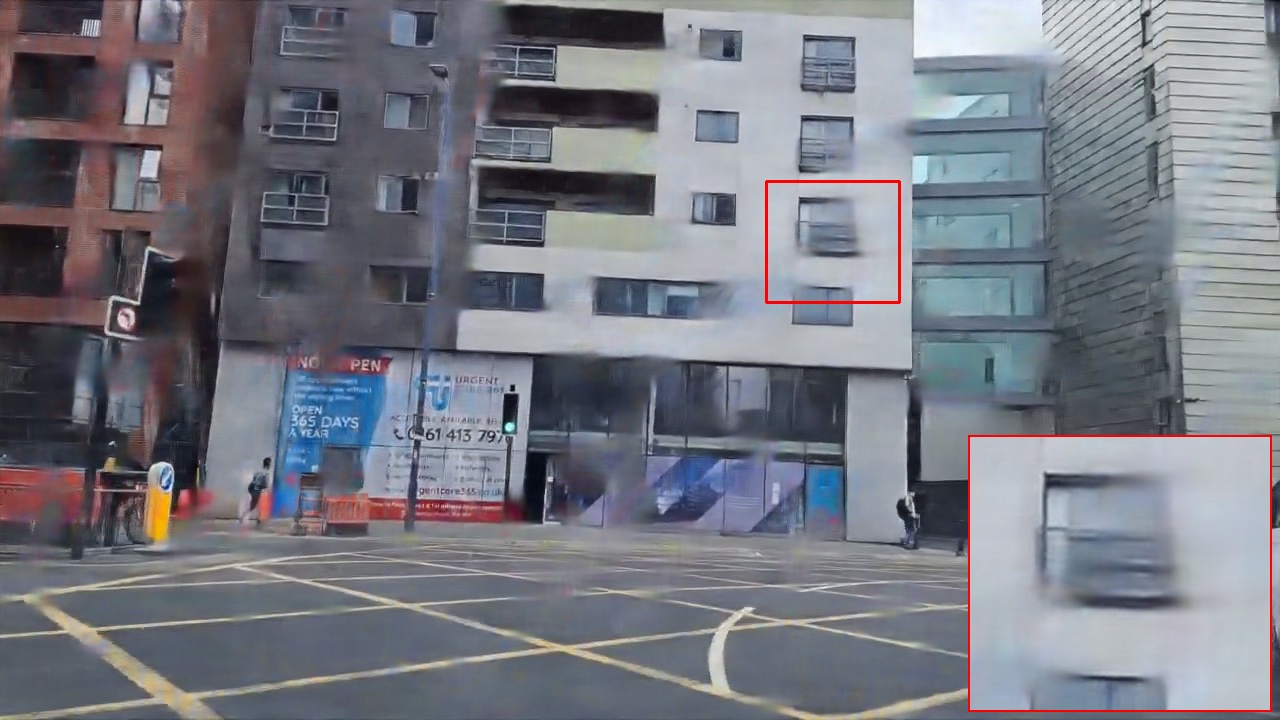} &
        \includegraphics[width=0.125\textwidth]{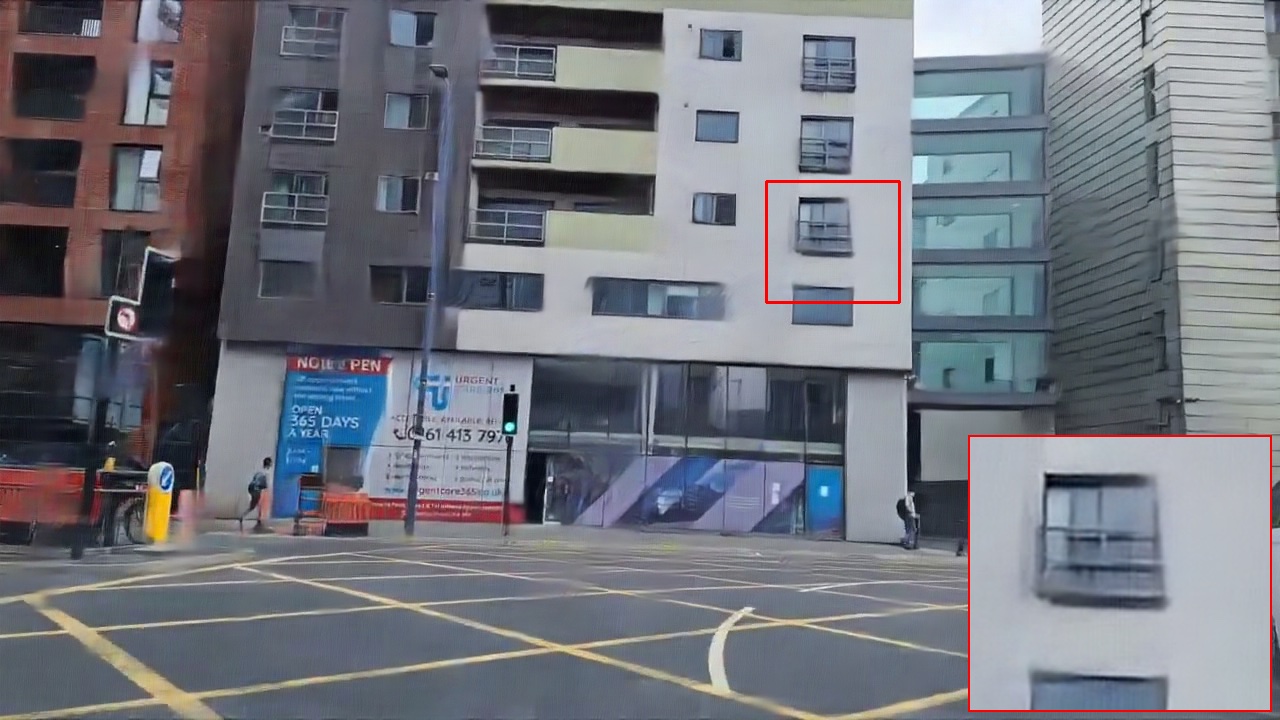} &
        \includegraphics[width=0.125\textwidth]{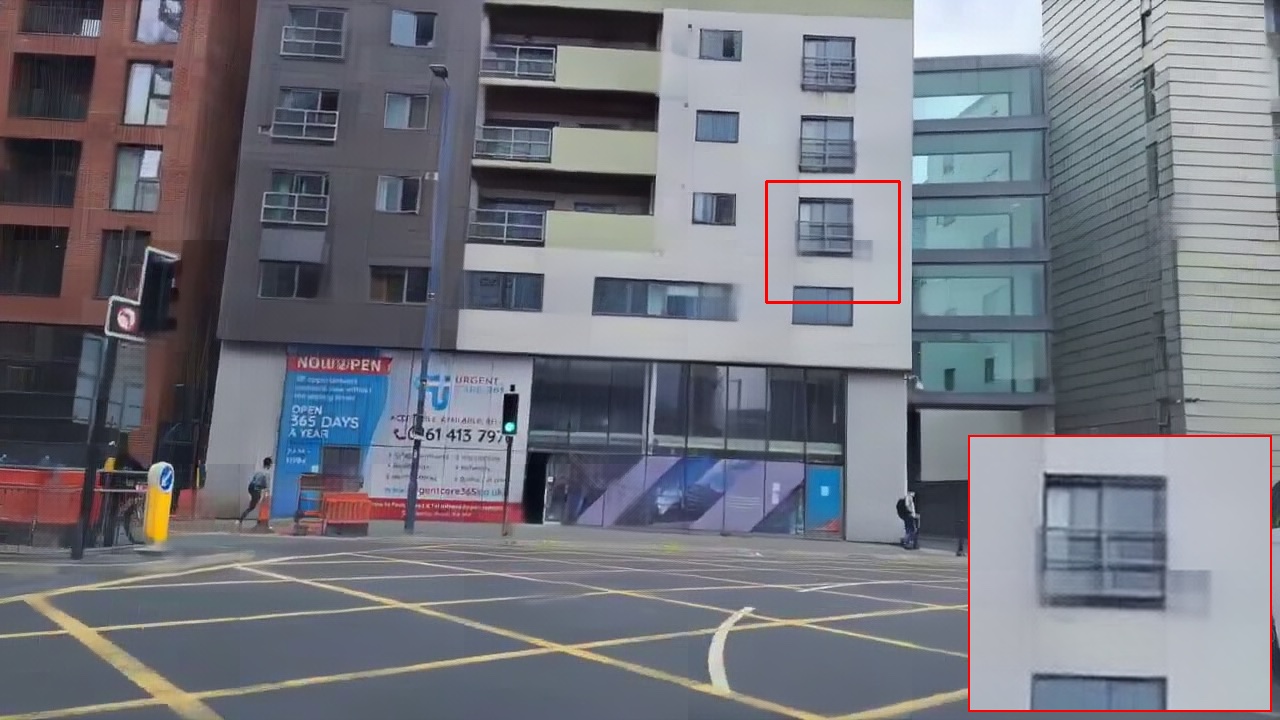} &
        \includegraphics[width=0.125\textwidth]{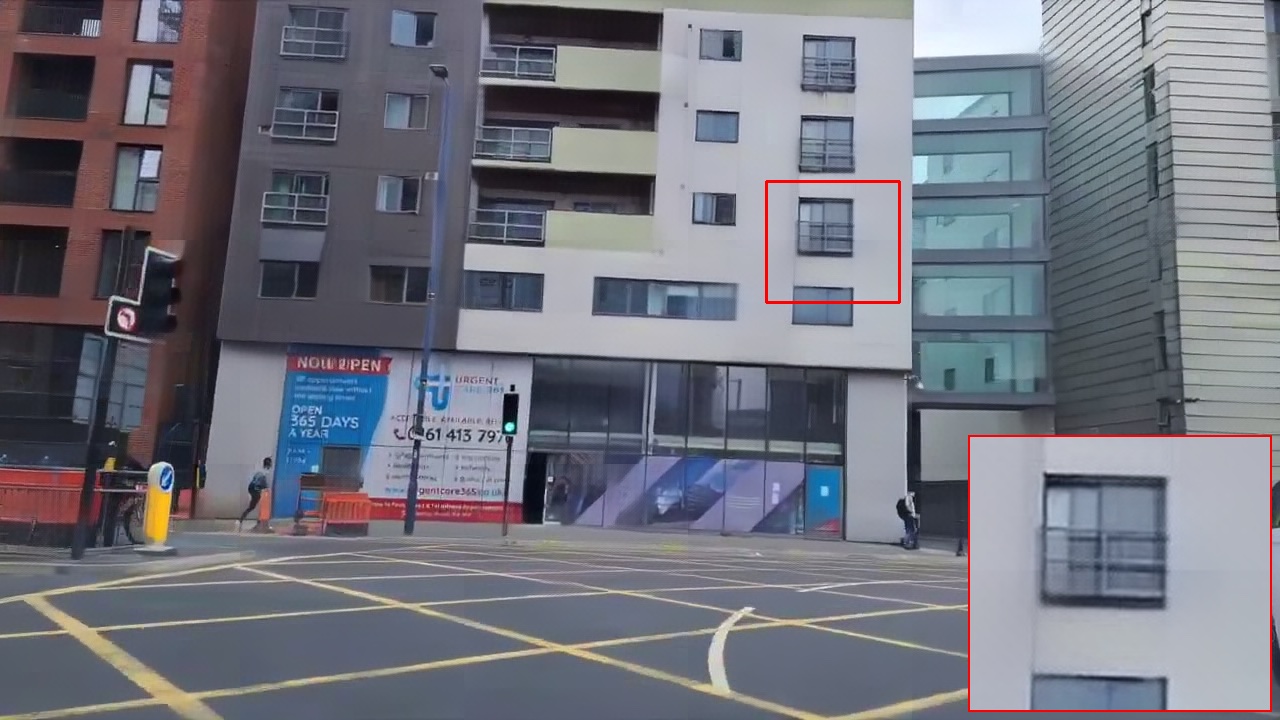} &
        \includegraphics[width=0.125\textwidth]{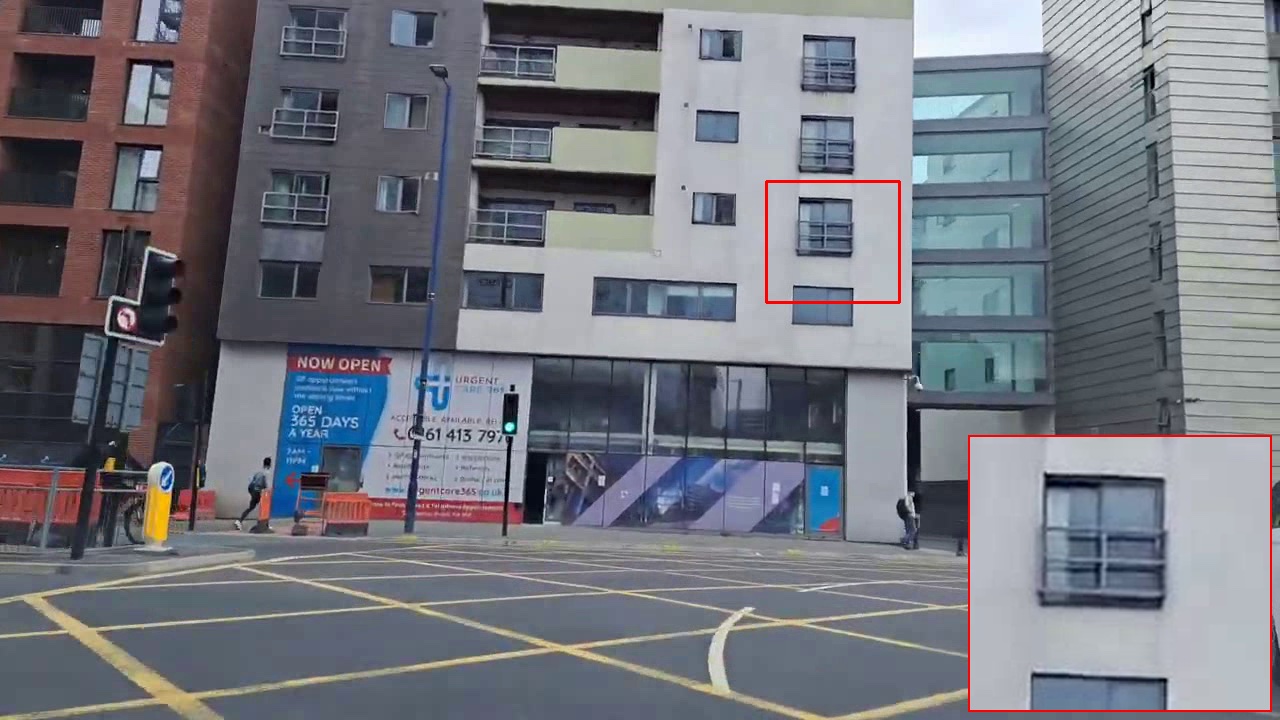} \\[2pt]

        \parbox[t]{0.125\textwidth}{\centering Input} &
        \parbox[t]{0.125\textwidth}{\centering DRSformer~\cite{zhang2022enhanced}} &
        \parbox[t]{0.125\textwidth}{\centering ViMP-Net~\cite{wu2023mask}} &
        \parbox[t]{0.125\textwidth}{\centering VDMamba~\cite{sun2025semi}} &
        \parbox[t]{0.125\textwidth}{\centering DeLiVR~\cite{sun2025delivr}} &
        \parbox[t]{0.125\textwidth}{\centering RainMamba~\cite{wu2024rainmamba}} &
        \parbox[t]{0.125\textwidth}{\centering Ours} &
        \parbox[t]{0.125\textwidth}{\centering GT}
    \end{tabular}


    \textbf{(b) Visual comparison on the RainVID\&SS dataset~\cite{sun2023event}.}\\[2pt]

    \begin{tabular}{@{}*{8}{c}@{}}
        \includegraphics[width=0.125\textwidth]{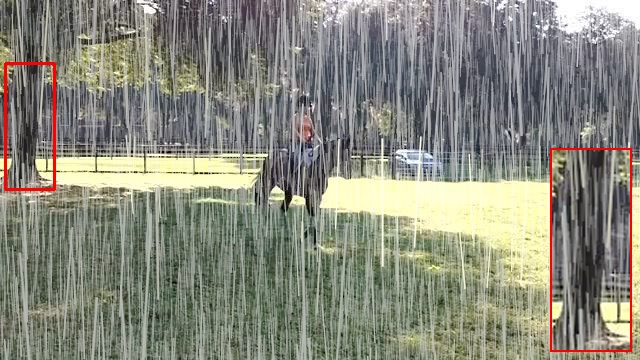} &
        \includegraphics[width=0.125\textwidth]{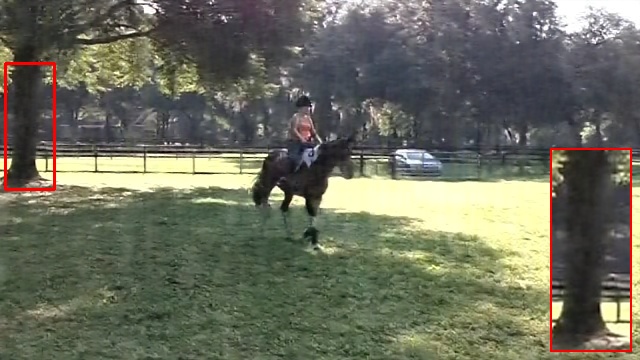} &
        \includegraphics[width=0.125\textwidth]{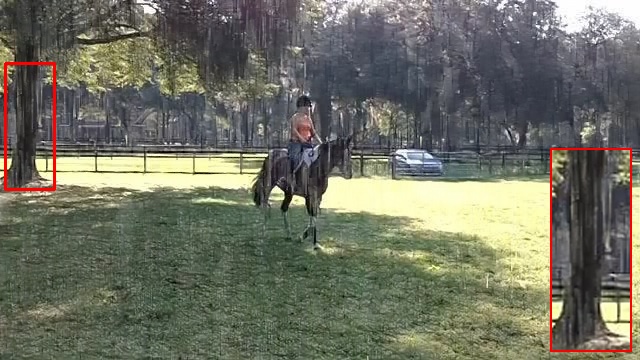} &
        \includegraphics[width=0.125\textwidth]{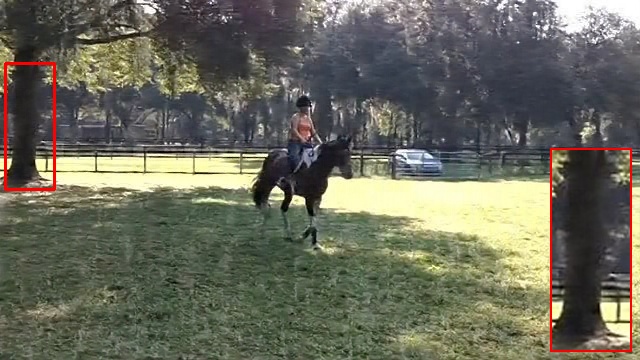} &
        \includegraphics[width=0.125\textwidth]{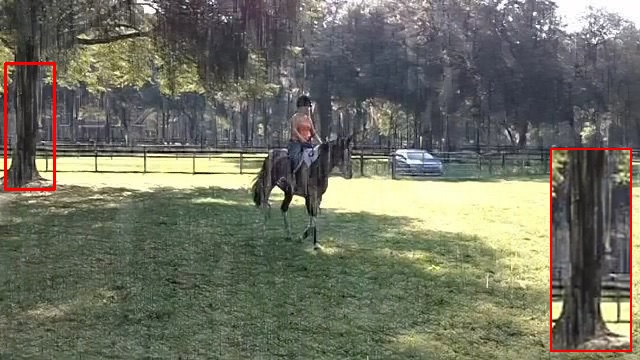} &
        \includegraphics[width=0.125\textwidth]{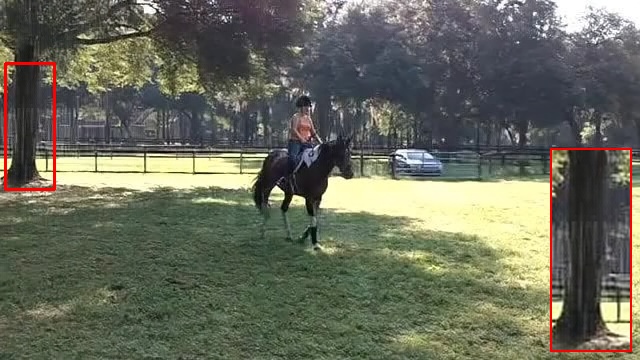} &
        \includegraphics[width=0.125\textwidth]{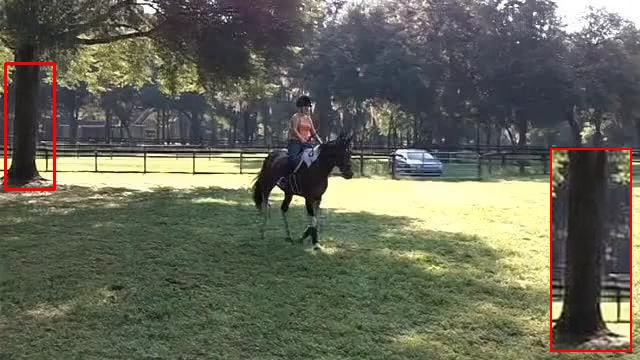} &
        \includegraphics[width=0.125\textwidth]{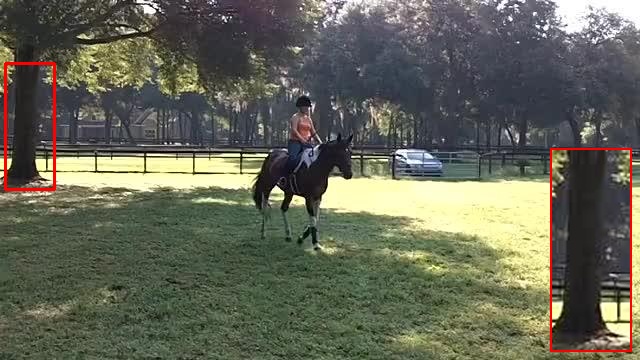} \\

        \includegraphics[width=0.125\textwidth]{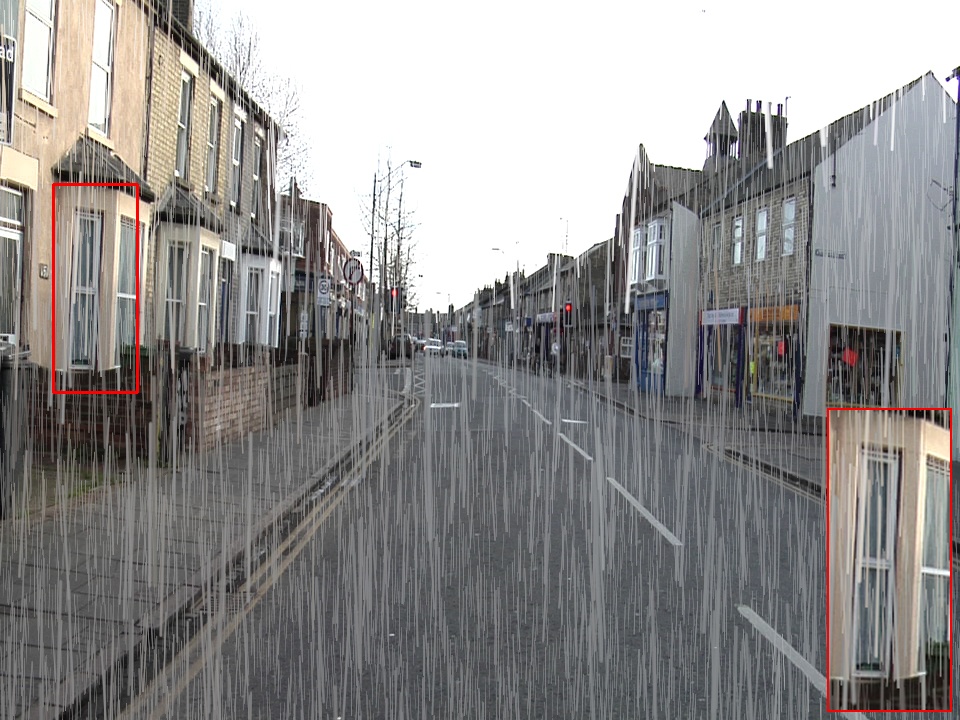} &
        \includegraphics[width=0.125\textwidth]{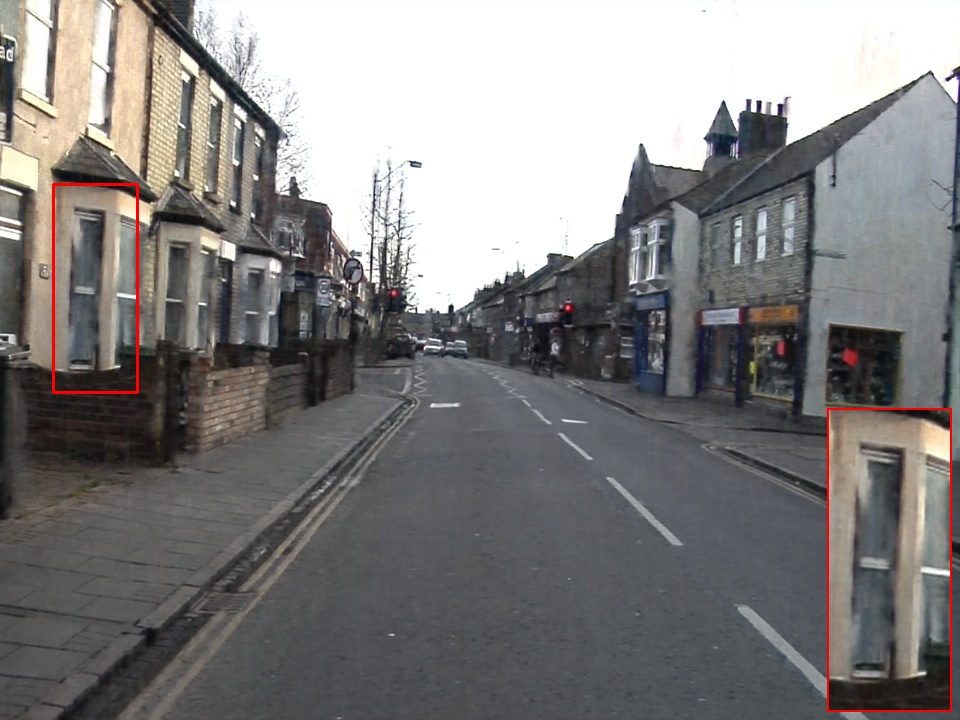} &
        \includegraphics[width=0.125\textwidth]{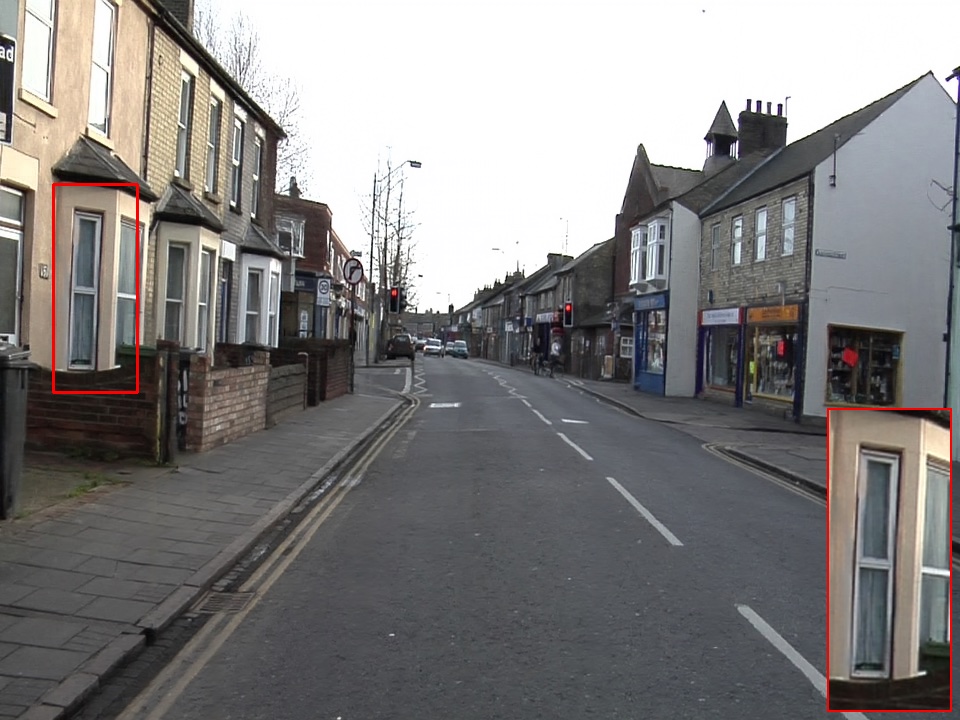} &
        \includegraphics[width=0.125\textwidth]{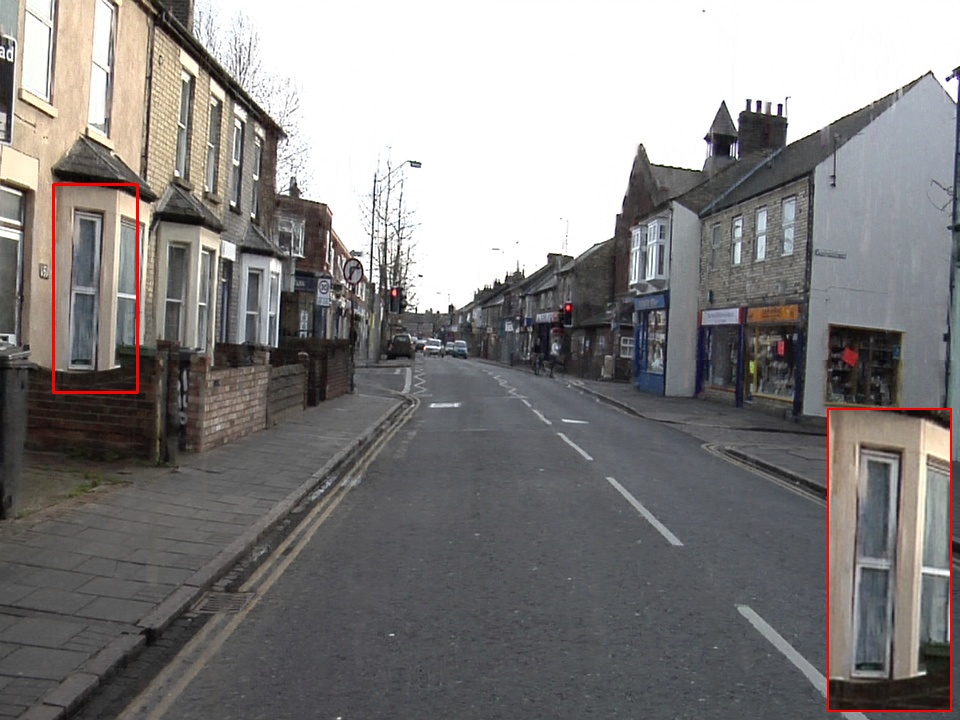} &
        \includegraphics[width=0.125\textwidth]{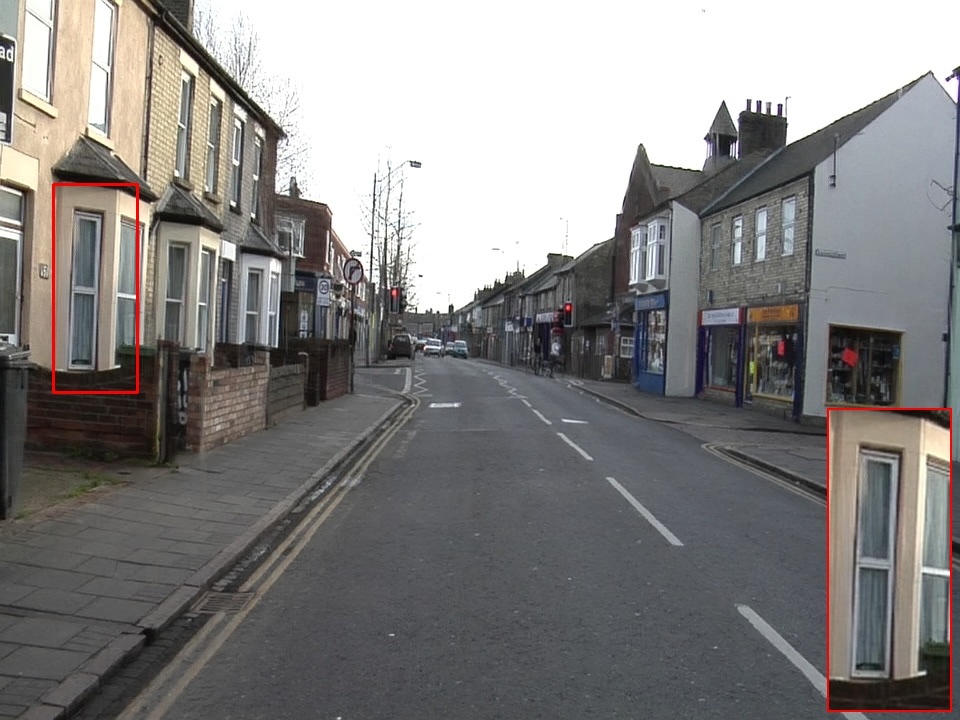} &
        \includegraphics[width=0.125\textwidth]{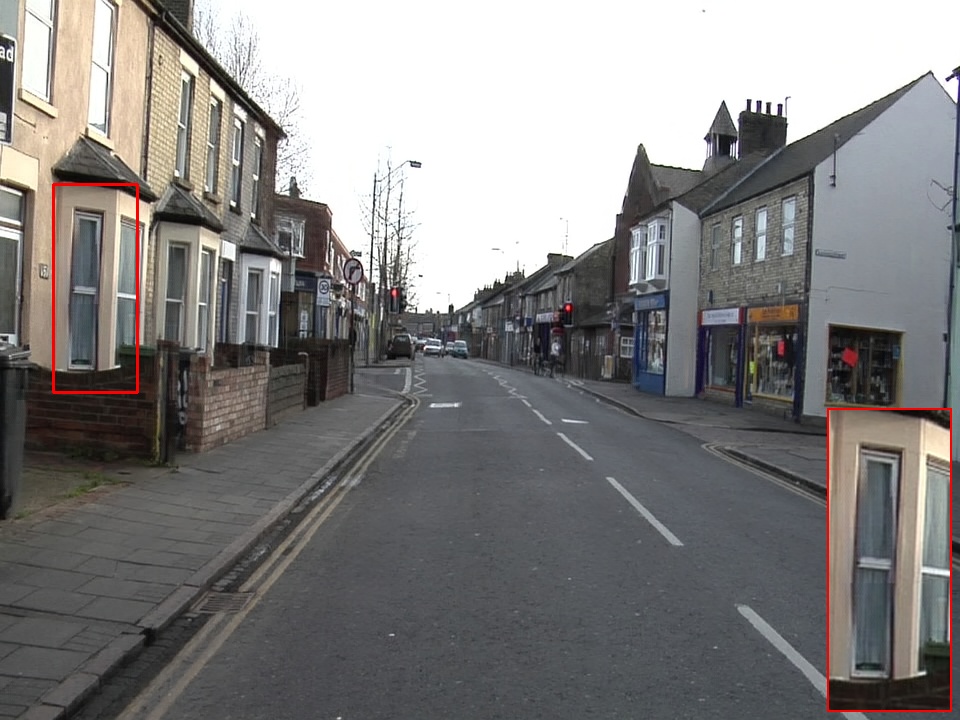} &
        \includegraphics[width=0.125\textwidth]{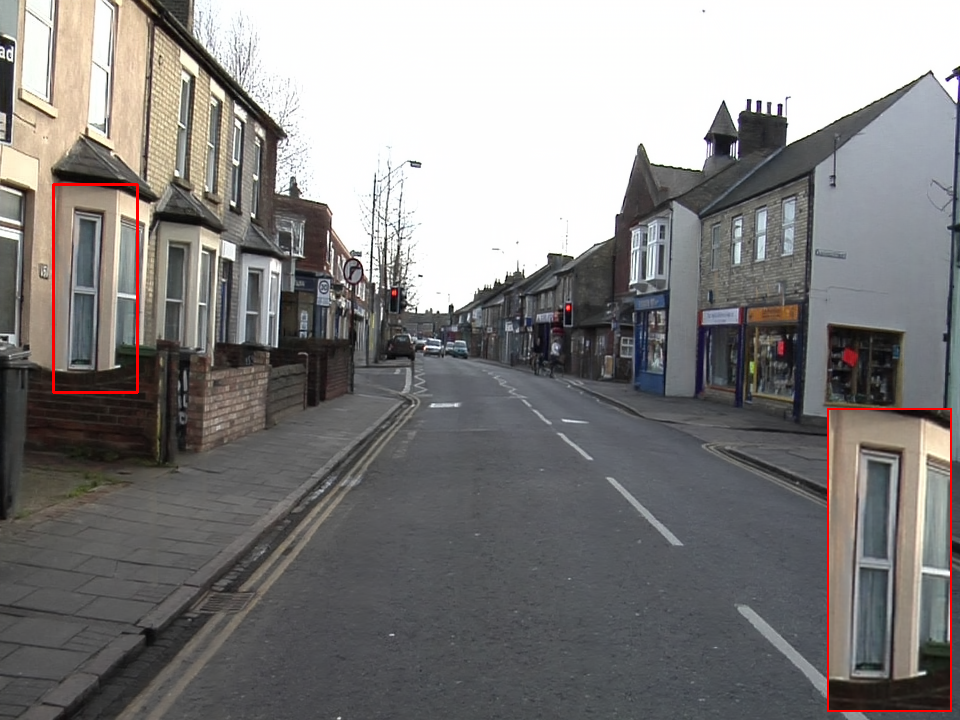} &
        \includegraphics[width=0.125\textwidth]{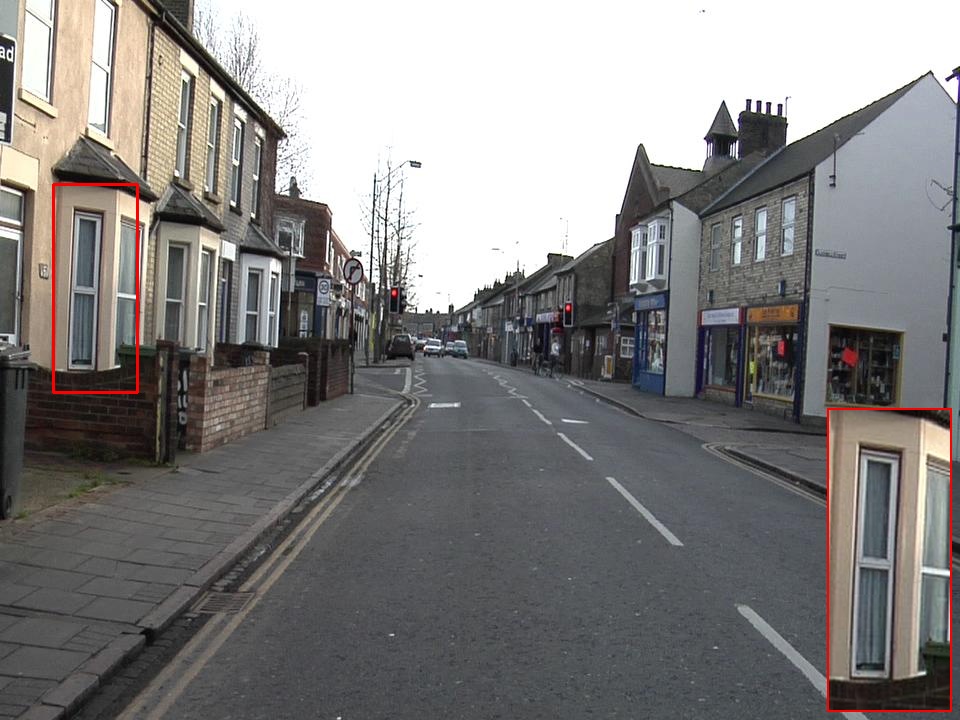} \\[2pt]

        \parbox[t]{0.125\textwidth}{\centering Input} &
        \parbox[t]{0.125\textwidth}{\centering S2VD~\cite{yue2021semi}} &
        \parbox[t]{0.125\textwidth}{\centering VDMamba~\cite{sun2025semi}} &
        \parbox[t]{0.125\textwidth}{\centering MPEVNet~\cite{sun2023event}} &
        \parbox[t]{0.125\textwidth}{\centering DeLiVR~\cite{sun2025delivr}} &
        \parbox[t]{0.125\textwidth}{\centering RainMamba~\cite{wu2024rainmamba}} &
        \parbox[t]{0.125\textwidth}{\centering Ours} &
        \parbox[t]{0.125\textwidth}{\centering GT}
    \end{tabular}


    \textbf{(c) Visual comparison on the RainSynAll100 dataset~\cite{yang2021recurrent}.}\\[2pt]

    \begin{tabular}{@{}*{8}{c}@{}}
        \includegraphics[width=0.125\textwidth]{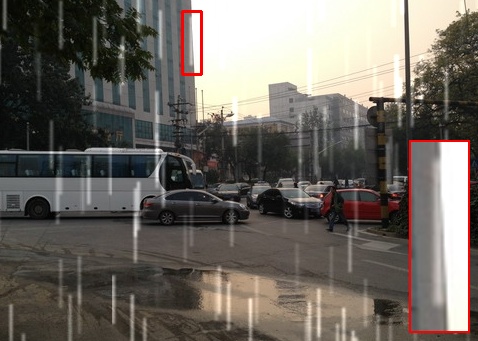} &
        \includegraphics[width=0.125\textwidth]{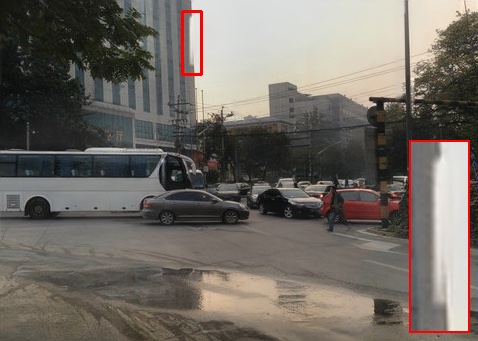} &
        \includegraphics[width=0.125\textwidth]{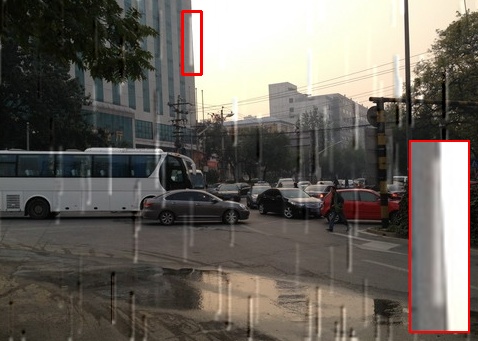} &
        \includegraphics[width=0.125\textwidth]{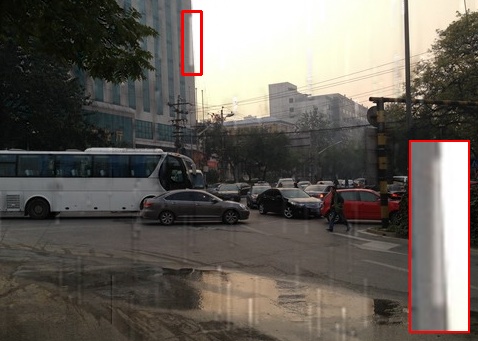} &
        \includegraphics[width=0.125\textwidth]{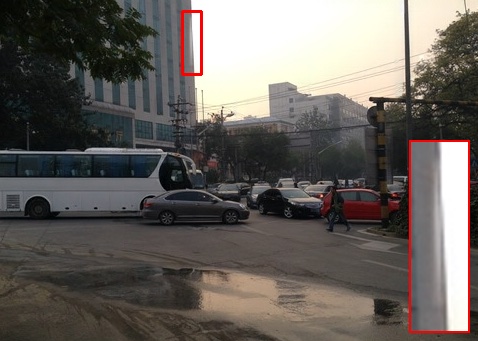} &
        \includegraphics[width=0.125\textwidth]{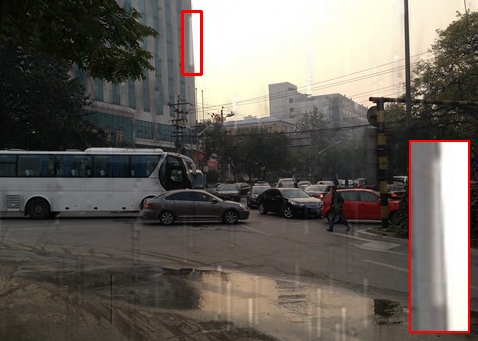} &
        \includegraphics[width=0.125\textwidth]{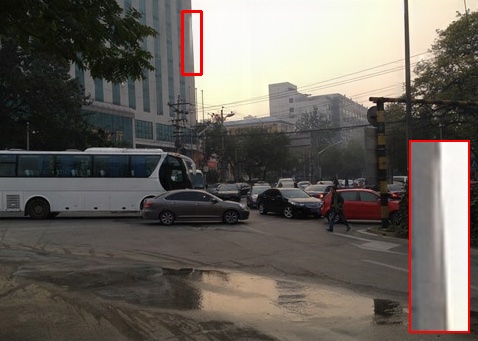} &
        \includegraphics[width=0.125\textwidth]{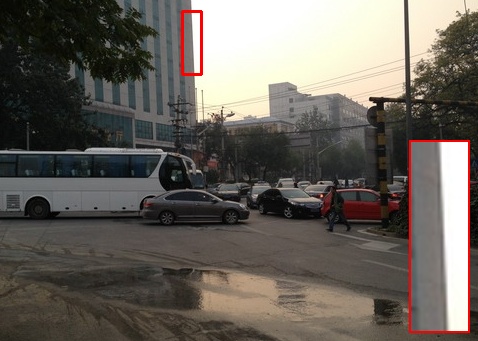} \\

        \includegraphics[width=0.125\textwidth]{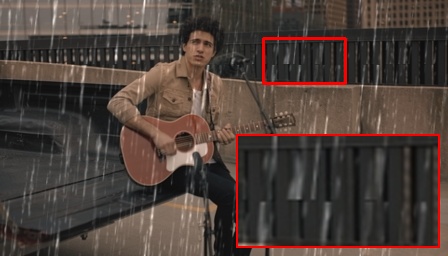} &
        \includegraphics[width=0.125\textwidth]{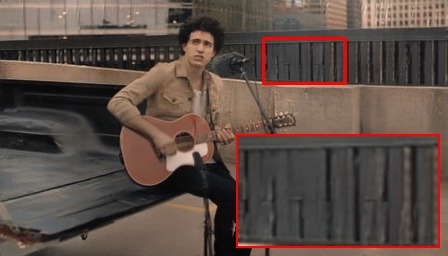} &
        \includegraphics[width=0.125\textwidth]{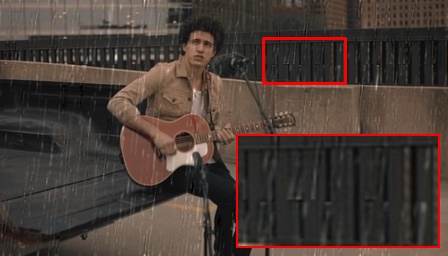} &
        \includegraphics[width=0.125\textwidth]{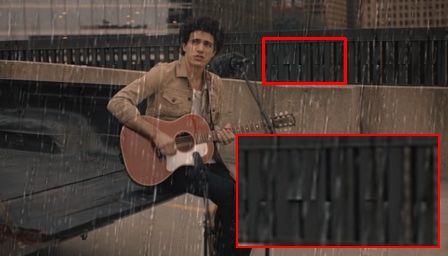} &
        \includegraphics[width=0.125\textwidth]{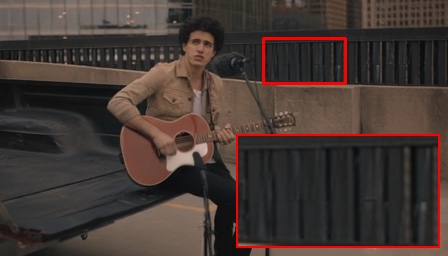} &
        \includegraphics[width=0.125\textwidth]{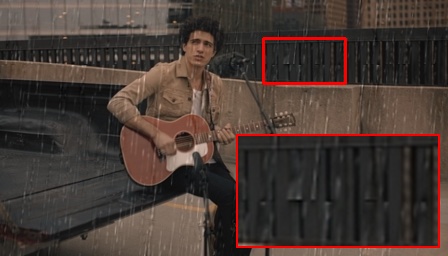} &
        \includegraphics[width=0.125\textwidth]{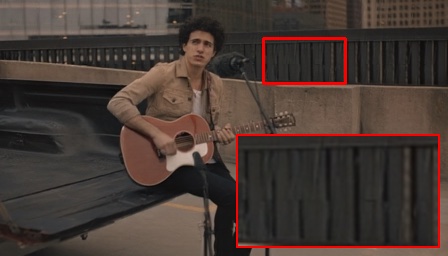} &
        \includegraphics[width=0.125\textwidth]{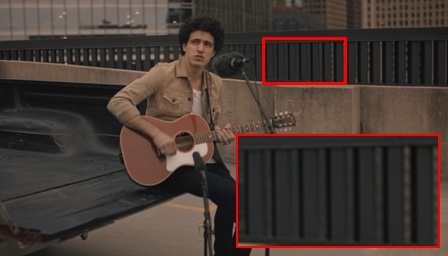} \\[2pt]

        \parbox[t]{0.125\textwidth}{\centering Input} &
        \parbox[t]{0.125\textwidth}{\centering RMFD~\cite{yang2021recurrent}} &
        \parbox[t]{0.125\textwidth}{\centering BasicVSR++~\cite{chan2022basicvsr++}} &
        \parbox[t]{0.125\textwidth}{\centering VDMamba~\cite{sun2025semi}} &
        \parbox[t]{0.125\textwidth}{\centering RainMamba~\cite{wu2024rainmamba}} &
        \parbox[t]{0.125\textwidth}{\centering DeLiVR~\cite{sun2025delivr}} &
        \parbox[t]{0.125\textwidth}{\centering Ours} &
        \parbox[t]{0.125\textwidth}{\centering GT}
    \end{tabular}


    \textbf{(d) Visual comparison on the LWDDS dataset~\cite{wen2023video}.}\\[2pt]

    \begin{tabular}{@{}*{8}{c}@{}}
        \includegraphics[width=0.125\textwidth]{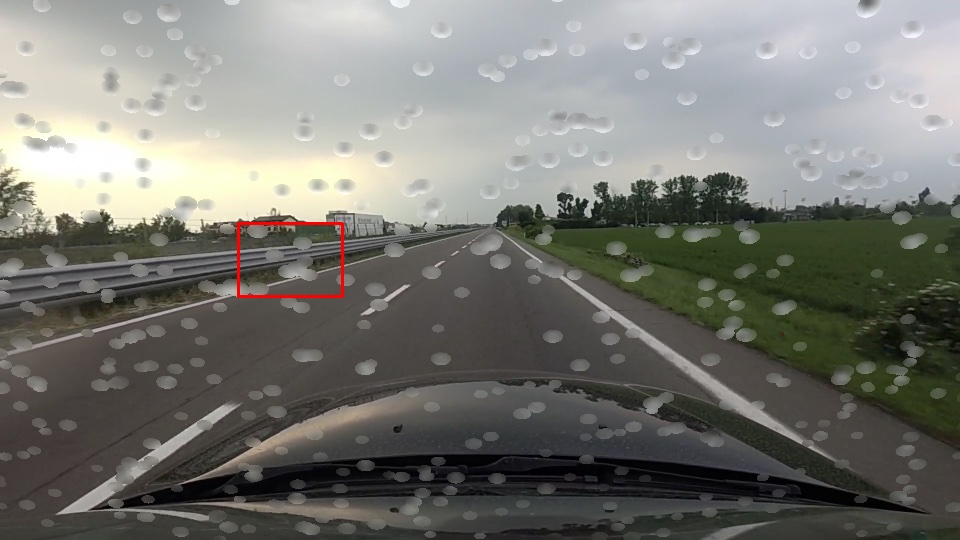} &
        \includegraphics[width=0.125\textwidth]{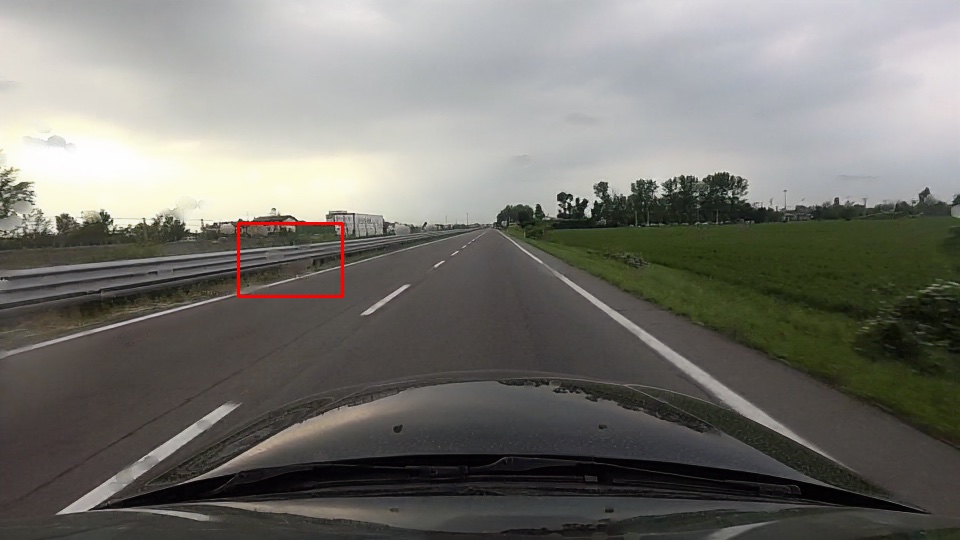} &
        \includegraphics[width=0.125\textwidth]{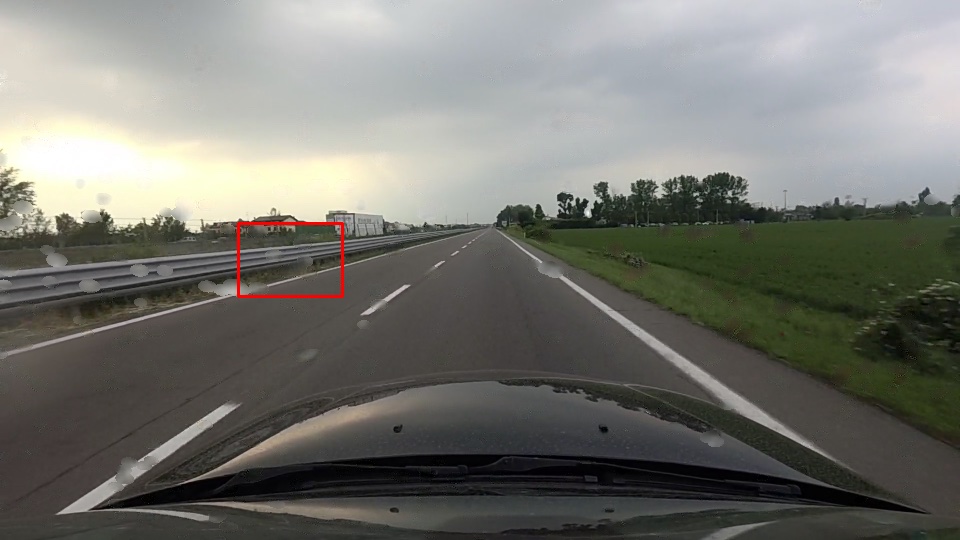} &
        \includegraphics[width=0.125\textwidth]{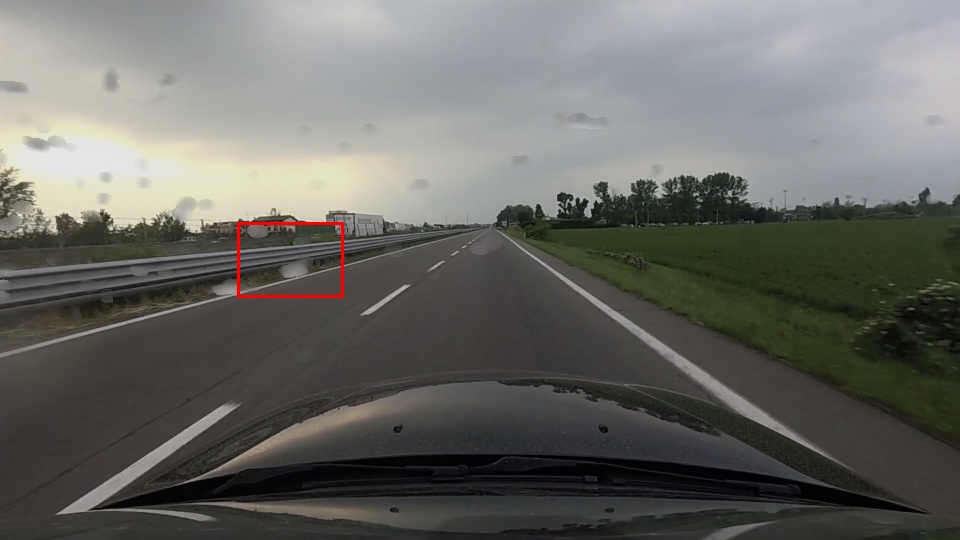} &
        \includegraphics[width=0.125\textwidth]{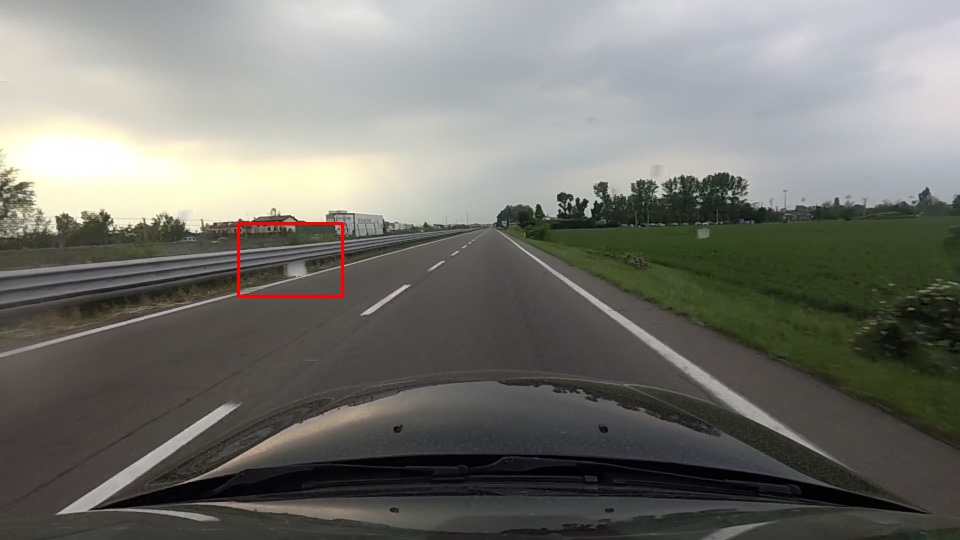} &
        \includegraphics[width=0.125\textwidth]{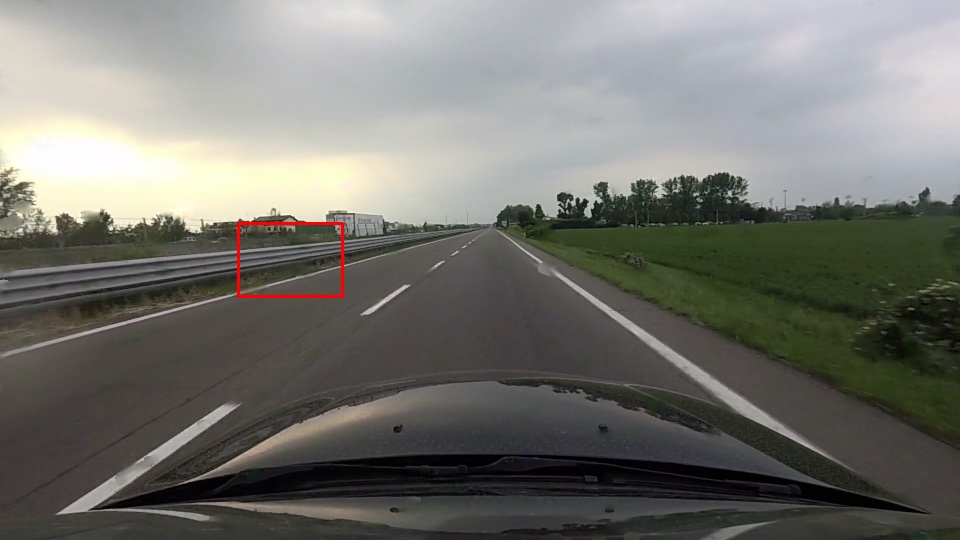} &
        \includegraphics[width=0.125\textwidth]{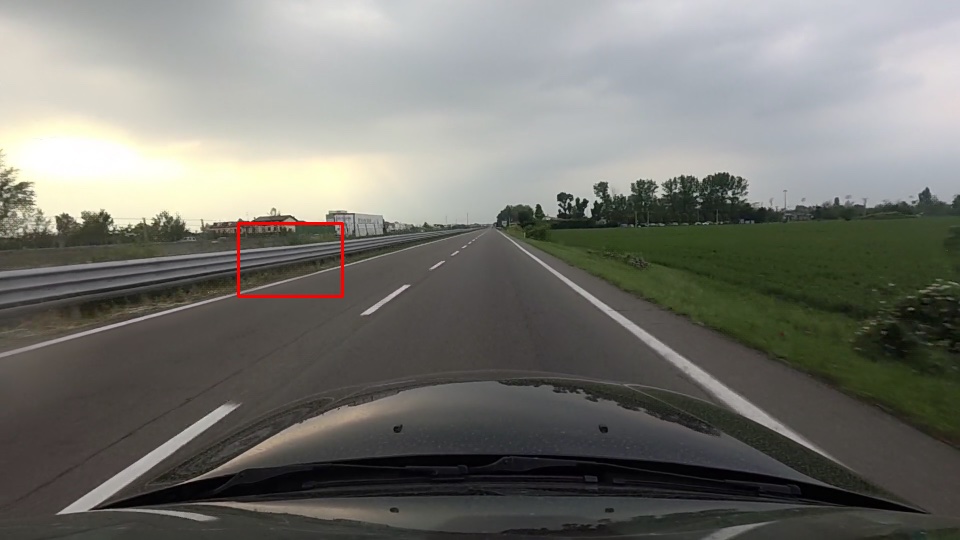} &
        \includegraphics[width=0.125\textwidth]{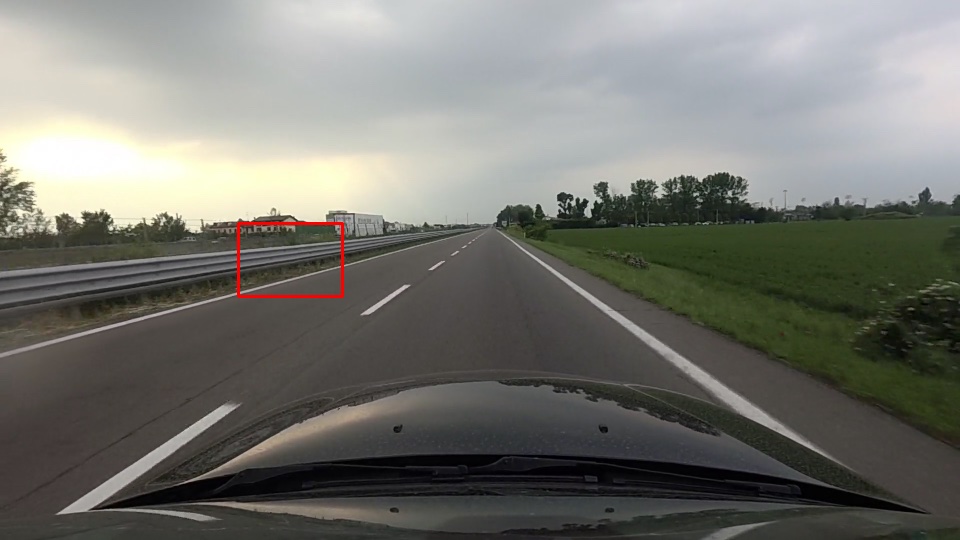}\\

        \includegraphics[width=0.125\textwidth]{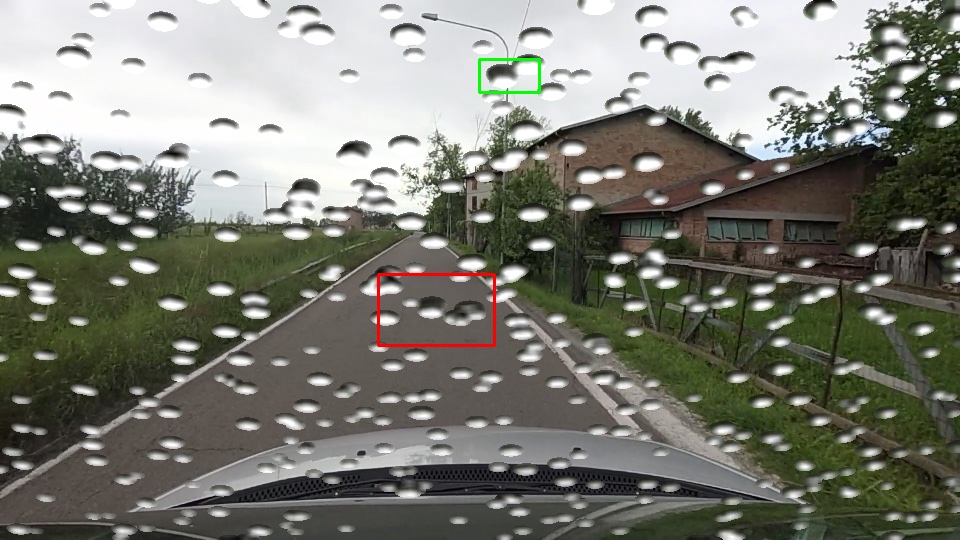} &
        \includegraphics[width=0.125\textwidth]{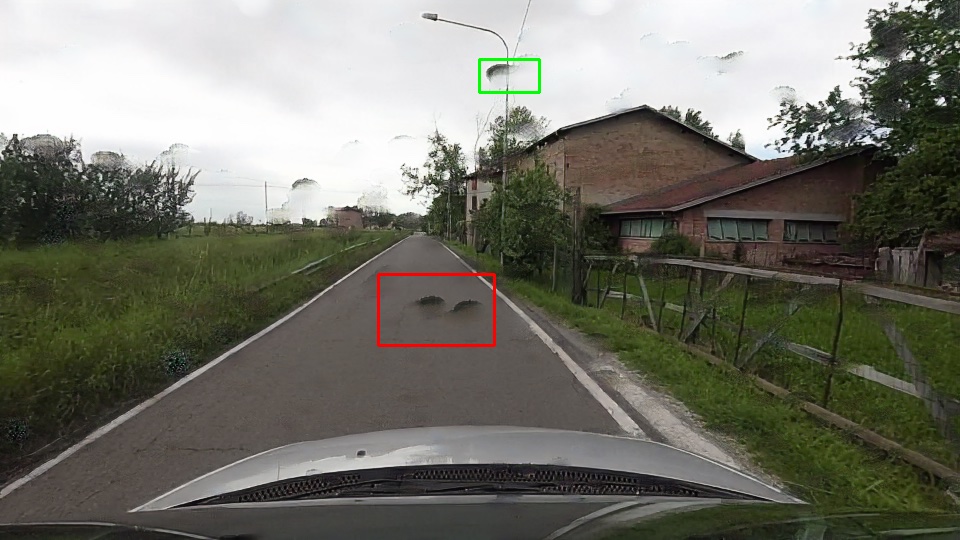} &
        \includegraphics[width=0.125\textwidth]{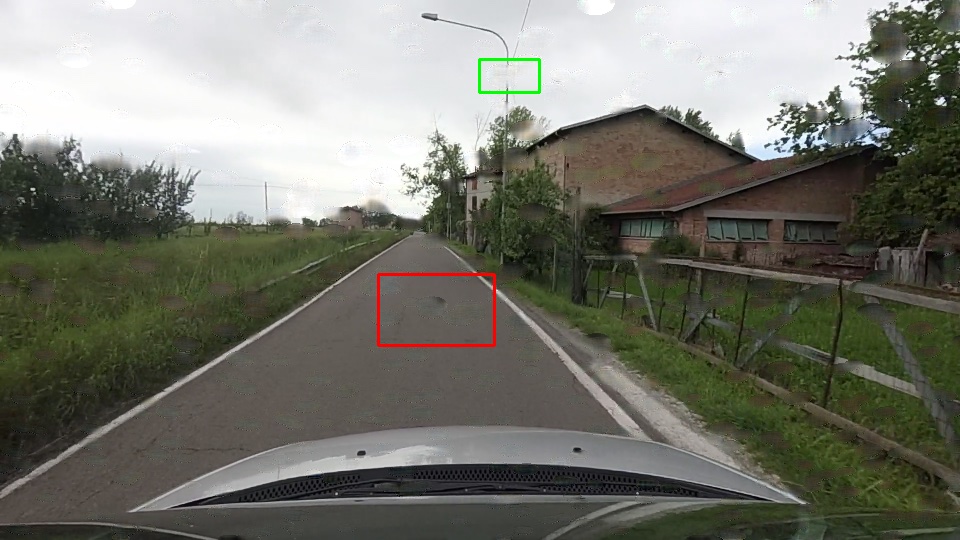} &
        \includegraphics[width=0.125\textwidth]{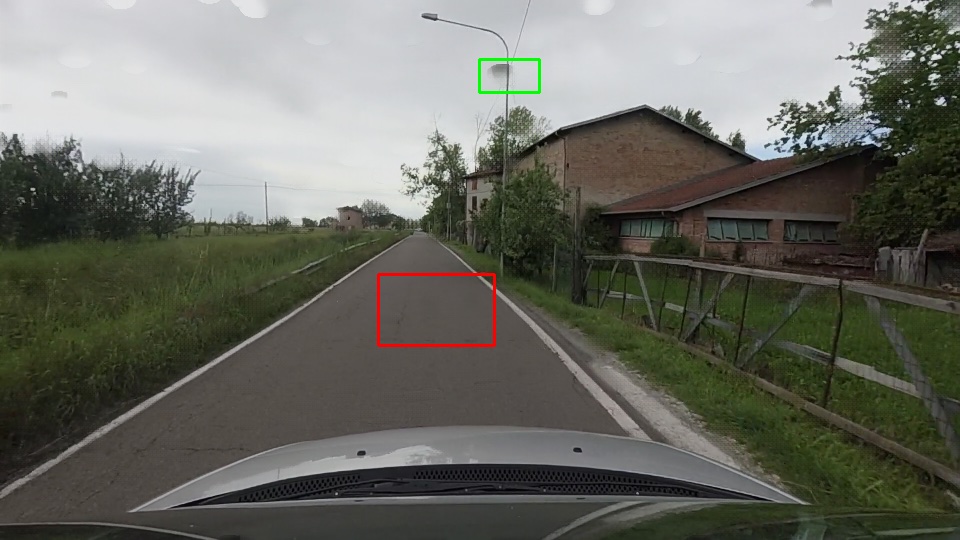} &
        \includegraphics[width=0.125\textwidth]{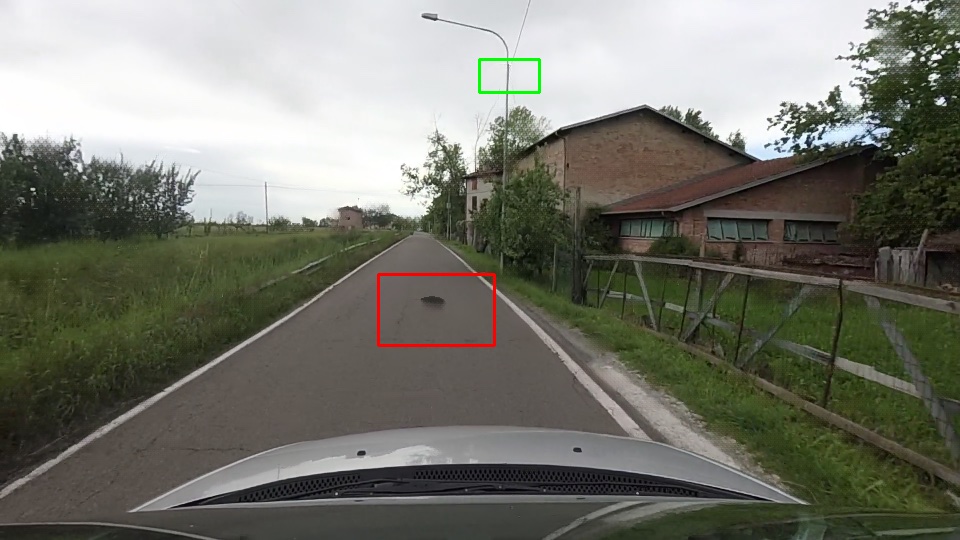} &
        \includegraphics[width=0.125\textwidth]{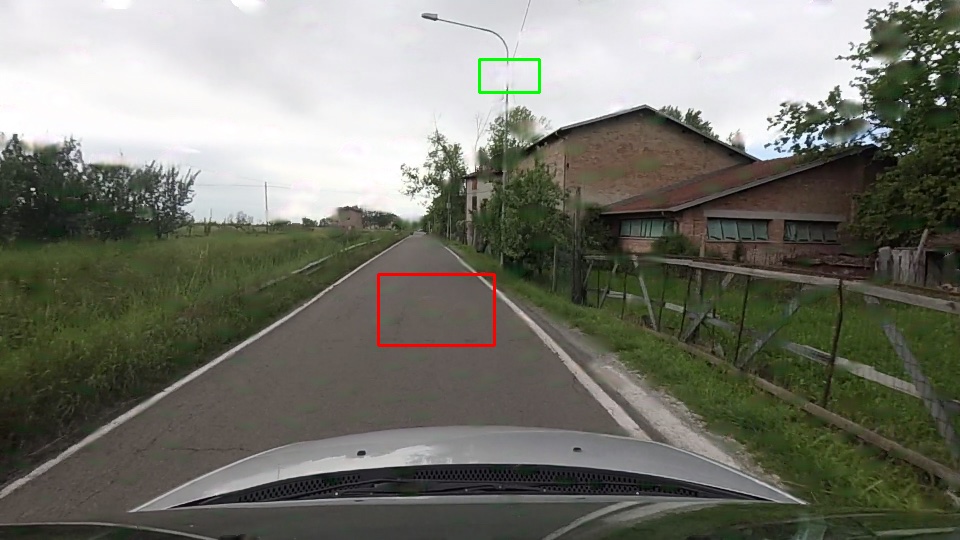} &
        \includegraphics[width=0.125\textwidth]{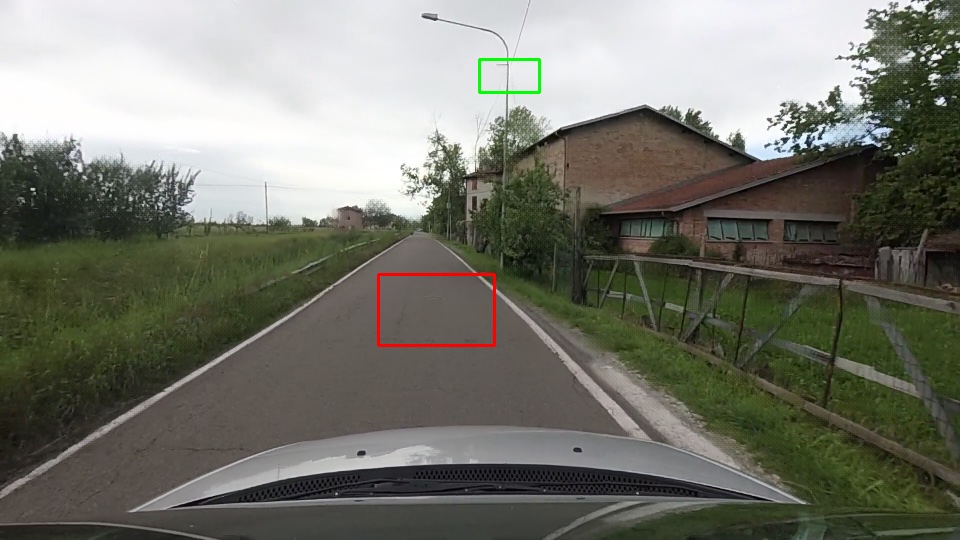} &
        \includegraphics[width=0.125\textwidth]{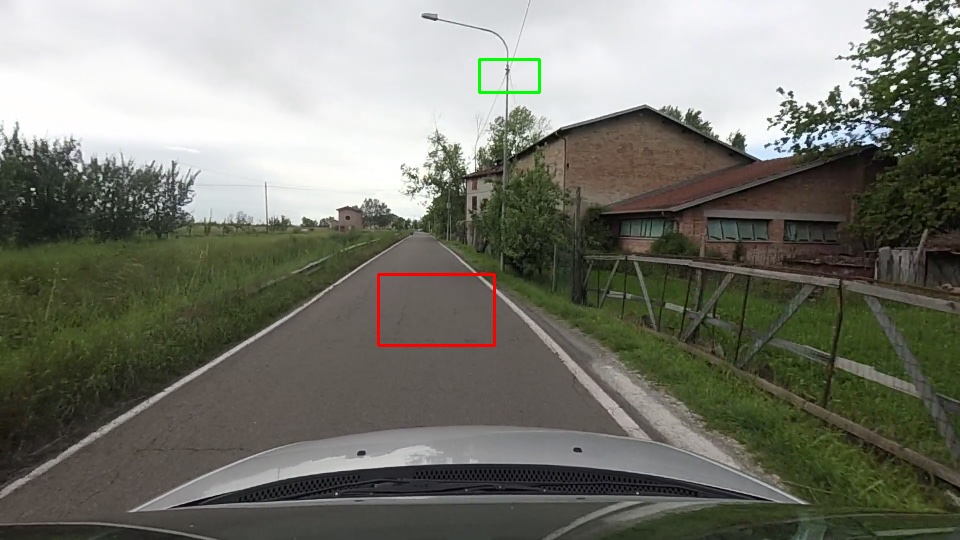}\\[2pt]

        \parbox[t]{0.125\textwidth}{\centering Input} &
        \parbox[t]{0.125\textwidth}{\centering VWR~\cite{wen2023video}} &
        \parbox[t]{0.125\textwidth}{\centering BasicVSR++~\cite{chan2022basicvsr++}} &
        \parbox[t]{0.125\textwidth}{\centering DeLiVR~\cite{sun2025delivr}} &
        \parbox[t]{0.125\textwidth}{\centering RainMamba~\cite{wu2024rainmamba}} &
        \parbox[t]{0.125\textwidth}{\centering VDMamba~\cite{sun2025semi}} &
        \parbox[t]{0.125\textwidth}{\centering Ours} &
        \parbox[t]{0.125\textwidth}{\centering GT}
    \end{tabular}

    \caption{Visual comparison of derained results produced by S3VD and state-of-the-art methods on the VRDS~\cite{wu2023mask}, RainVID\&SS~\cite{sun2023event}, RainSynAll100~\cite{yang2021recurrent}, and LWDDS~\cite{wen2023video} datasets. Please zoom in for a better illustration.}
    \label{fig:visual_comparison_all}
    \label{fig:VRDS}
    \label{fig:RainVIDSS}
    \label{fig:RainSynAll100}
    \label{fig:LWDDS}
\end{figure*}

\noindent \textbf{Experimental Setup.}
\label{sec:ex}
Our network is trained on two NVIDIA RTX 3090 GPUs based on the PyTorch platform. We perform separate training and testing procedures for different datasets. For the VRDS~\cite{wu2023mask} and LWDDS~\cite{wen2023video} datasets, input frames are randomly cropped to a spatial resolution of $256\times256$, with each video clip containing 5 frames. We employ the Adam optimizer with a 1.0 polynomial scheduler, setting the initial learning rate to $4\times 10^{-4}$. Training is conducted with a batch size of 4, including 2k warm-up iterations and a total of 300k iterations.
For the RainVID\&SS~\cite{sun2023event} and RainSynAll100~\cite{yang2021recurrent}, input frames are randomly cropped to $128\times128$ resolution, with 7 and 5 frames per video clip, respectively. In this case, the initial learning rate is adjusted to $2\times 10^{-4}$. Throughout all experiments, we utilize a ConvNeXt backbone pretrained on ImageNet as the encoder and the plainest version of DINOv2 (\text{dinov2\_vits14}). The STSF modules were configured with $N_1=2$, $N_2=3$, and $N_3=2$. 

During inference, an arbitrary-length video is processed using a centered sliding window with stride one and the same fixed window size as in training ($K=5$ for VRDS, LWDDS, and RainSynAll100, and $K=7$ for RainVID\&SS). Boundary windows are formed by replicating the first or last valid frame, and only the restored center frame of each window is retained. Thus, a video containing $T$ frames produces $T$ restored frames.

\subsection{Comparisons with State-of-the-art Methods}
\noindent\textbf{Quantitative Comparisons.}
As demonstrated in Tables~\ref{tab:cmp}, our proposed S3VD framework consistently outperforms state-of-the-art methods across all datasets, highlighting its versatility in handling various rain conditions. 
For complex, 
S3VD shows exceptional ability in the challenging VRDS dataset,
surpassing RainMamba by 0.39 dB in PSNR in heavy rain scenarios. 

For rain streak removal, S3VD delivers remarkable performance across different datasets. On the RainVID\&SS dataset, our method achieves outstanding PSNR/SSIM values of 35.19 dB/0.9583, consistently outperforming previous methods regardless of the sequence length. This temporal robustness can be attributed to the modeling of long-range dependencies via STSF and robust semantic priors. 
Similarly, on the RainSynAll100 dataset, which focuses on synthetic rain streak patterns, our approach demonstrates substantial improvements, exceeding DeLiVR by 0.65 dB in PSNR and 0.0068 in SSIM. This significant performance leap derives from our novel DG-Mamba layer, which integrates spatio-temporal information for precise presentation of rain streaks. 

For raindrop removal, our framework achieves exceptional results in LWDDS dataset with superior performance, outperforming RainMamba by 0.74 dB. 
It is primarily attributed to our spatio-temporal modeling and robust semantic priors, which excel at reconstructing details with multi-scale semantic cues and spatio-temporal correlations exploration. 


\noindent\textbf{Visual Comparisons.}
Figs.~\ref{fig:visual_comparison_all} present visual comparisons of our S3VD and state-of-the-art methods in four rainy datasets. 
Our approach demonstrates superior performance in both rain streak and raindrop removal while retaining the major background structures and visible details. Taking the first sample in VRDS (Fig.~\ref{fig:VRDS}(a)) as an example, our S3VD method effectively captures and removes dense rain while maintaining the structural integrity of buildings and vehicles, while other methods struggle to produce satisfied deraining results with obvious rain residues or blurred details.

For rain streak removal (Figs.~\ref{fig:RainVIDSS}(b) and (c)), our approach shows significant advantages. In heavy rain scenarios, such as the first example from RainVID\&SS (Fig.~\ref{fig:RainVIDSS}(b)), the competing methods struggle to remove rain streaks effectively.
In contrast, our S3VD, guided by the robust semantic context of the MSSF module, removes a substantial portion of the visible rain streaks in this example while retaining the major structure and fine edges of the background tree.

In the task of raindrop removal, while other methods tend to leave behind obvious blurry artifacts or distortions where raindrops occluded the scene, our S3VD generates much cleaner and more plausible reconstructions (Figs.~\ref{fig:LWDDS}(d)). This is because our STSF module effectively performs spatio-temporal fusion to reconstruct the occluded regions. The DG-Mamba layer adaptively gathers and fuses information from adjacent frames, allowing it to reconstruct the scene behind the raindrops with high fidelity instead of just blurring the area.

\begin{figure*}[t]
    \centering
    \renewcommand\arraystretch{0.2}
    \setlength{\tabcolsep}{0pt} 

    \begin{tabular}{*{6}{c}} 

        \includegraphics[width=0.155\textwidth]{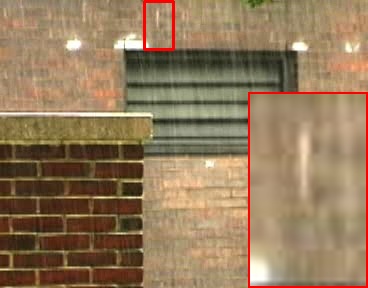} &
        \includegraphics[width=0.155\textwidth]{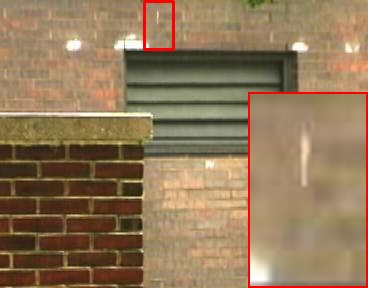} &
        \includegraphics[width=0.155\textwidth]{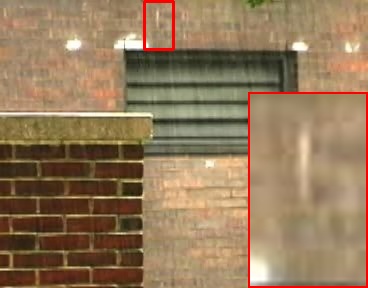} &
        \includegraphics[width=0.155\textwidth]{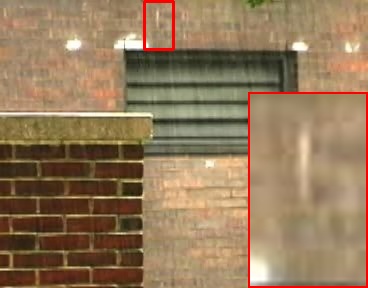} &
        \includegraphics[width=0.155\textwidth]{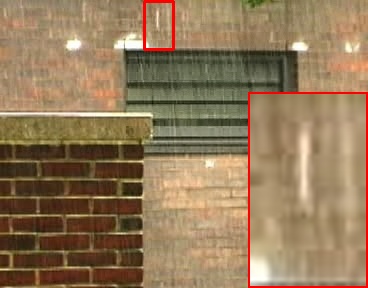} &
        \includegraphics[width=0.155\textwidth]{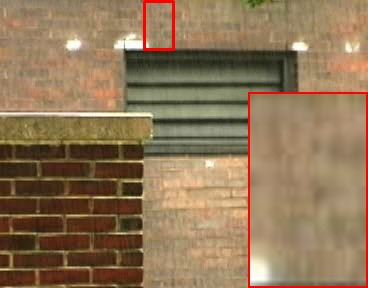} \\
        \includegraphics[width=0.155\textwidth]{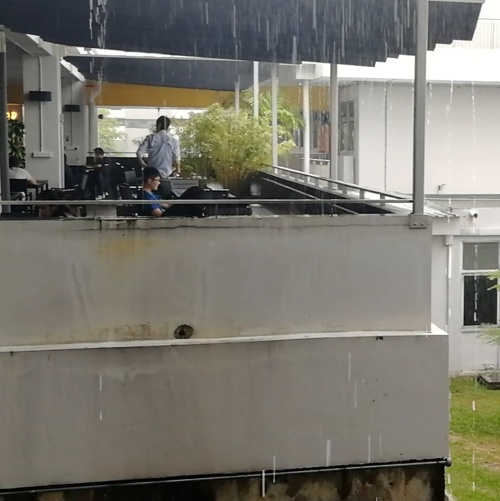} &
        \includegraphics[width=0.155\textwidth]{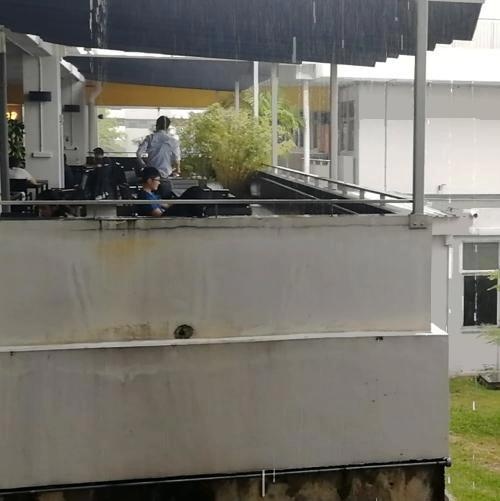} &
        \includegraphics[width=0.155\textwidth]{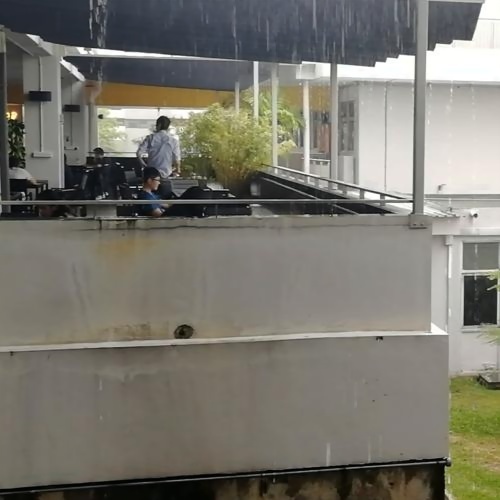} &
        \includegraphics[width=0.155\textwidth]{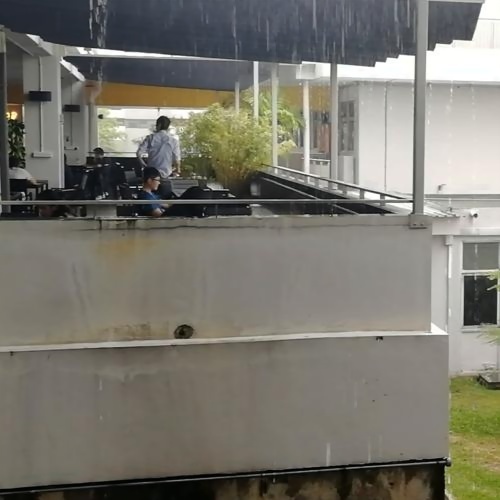} &
        \includegraphics[width=0.155\textwidth]{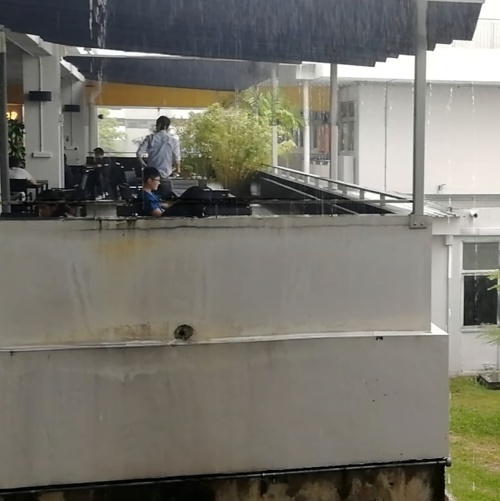} &
        \includegraphics[width=0.155\textwidth]{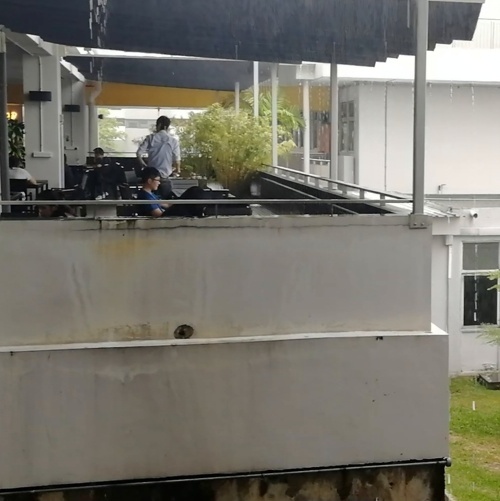} \\
        \includegraphics[width=0.155\textwidth]{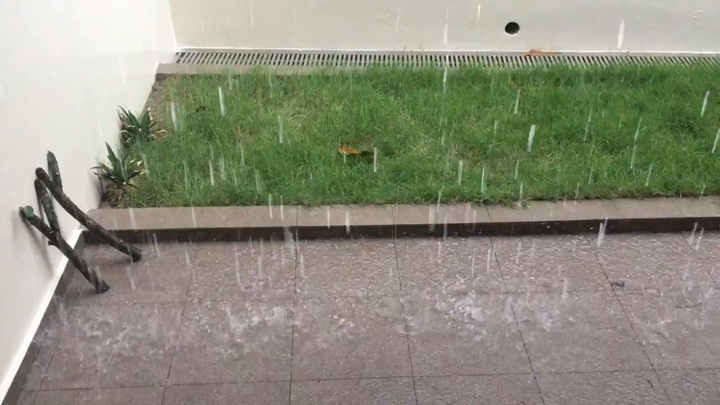} &
        \includegraphics[width=0.155\textwidth]{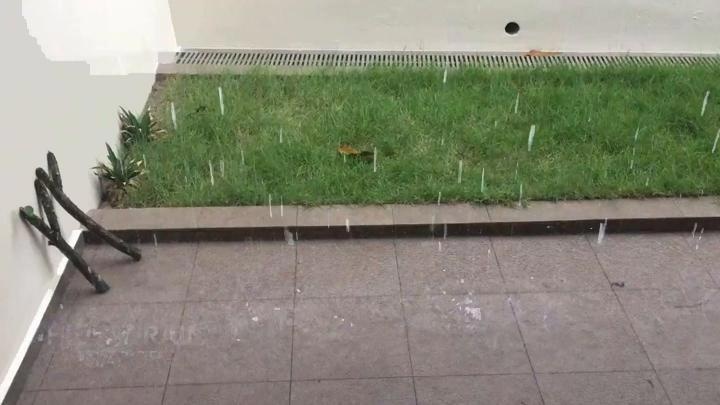} &
        \includegraphics[width=0.155\textwidth]{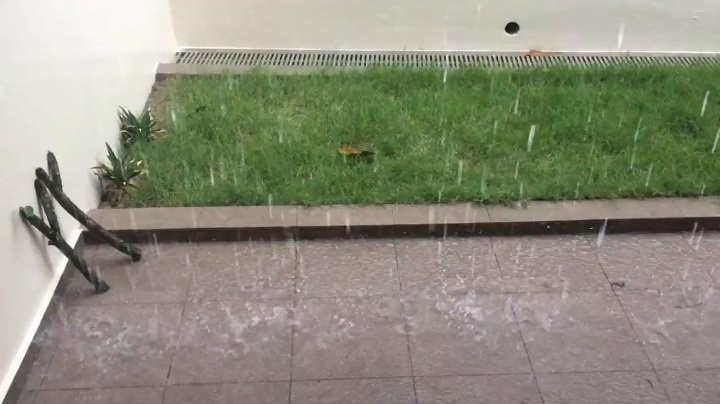} &
        \includegraphics[width=0.155\textwidth]{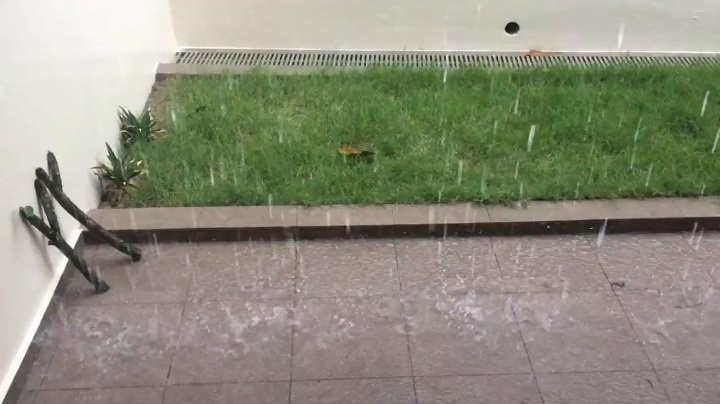} &
        \includegraphics[width=0.155\textwidth]{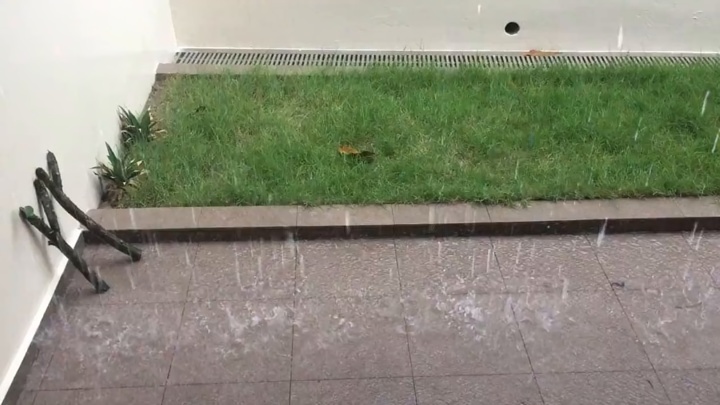} &
        \includegraphics[width=0.155\textwidth]{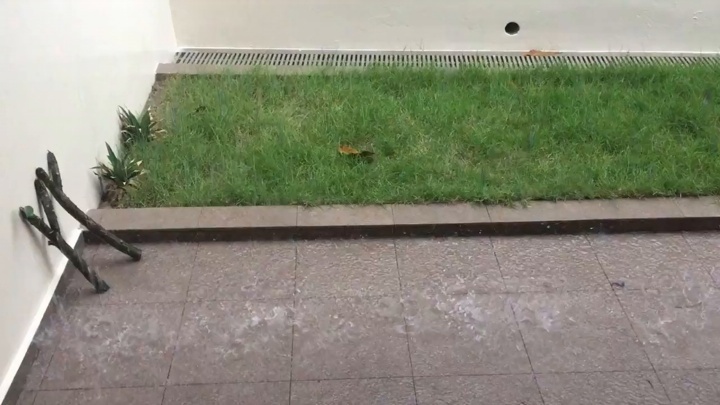} \\[3pt]
        
        \parbox[t]{0.155\textwidth}{\centering Input} &
        \parbox[t]{0.155\textwidth}{\centering MSCSC~\cite{li2018video}} &
        \parbox[t]{0.155\textwidth}{\centering RMFD~\cite{yang2021recurrent}} & 
        \parbox[t]{0.155\textwidth}{\centering S2VD~\cite{yue2021semi}} &
        \parbox[t]{0.155\textwidth}{\centering RainMamba~\cite{wu2024rainmamba}} & 
        \parbox[t]{0.155\textwidth}{\centering Ours} \\
        
    \end{tabular}
    
    \caption{Visual Comparison of derained results between our S3VD and other state-of-the-art methods on the real-world dataset from~\cite{sun2023event}. Please zoom in for a better illustration.}
    \label{fig:Real-World} 

\end{figure*}

\begin{figure*}[t]
    \centering
    \renewcommand\arraystretch{0.2}
    \setlength{\tabcolsep}{0pt} 

    \begin{tabular}{*{6}{c}} 

        \multicolumn{3}{c}{\textbf{(a) Clean (Ground Truth)}} & \multicolumn{3}{c}{\textbf{(b) Rainy}} \\[3pt]

        \includegraphics[width=0.155\textwidth]{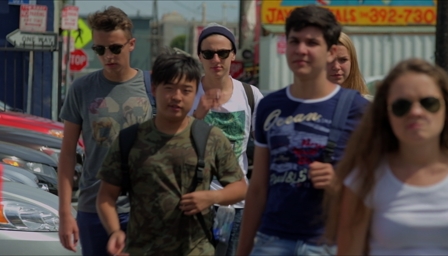} &
        \includegraphics[width=0.155\textwidth]{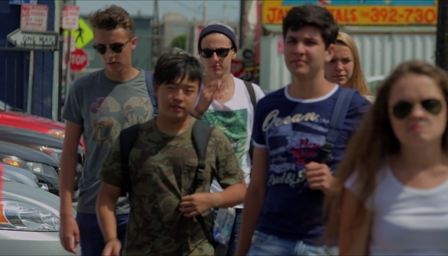} &
        \includegraphics[width=0.155\textwidth]{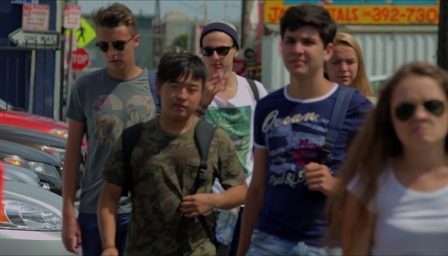} &
        \includegraphics[width=0.155\textwidth]{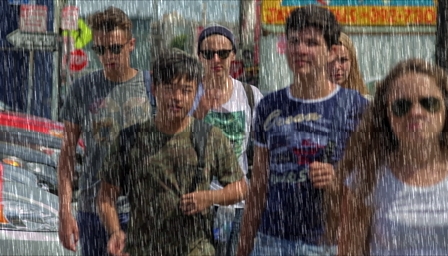} &
        \includegraphics[width=0.155\textwidth]{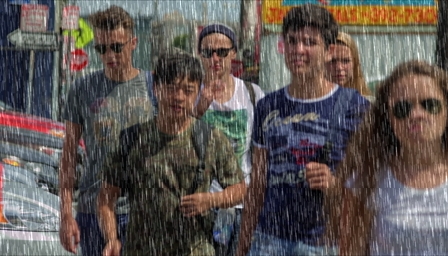} &
        \includegraphics[width=0.155\textwidth]{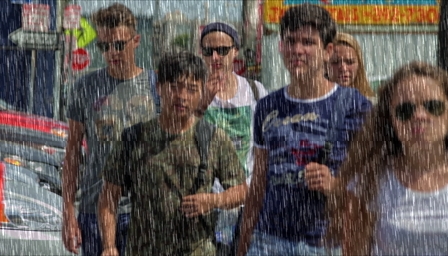} \\
        
        \includegraphics[width=0.155\textwidth]{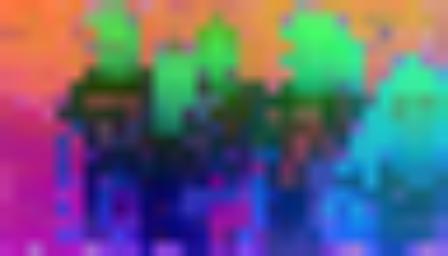} &
        \includegraphics[width=0.155\textwidth]{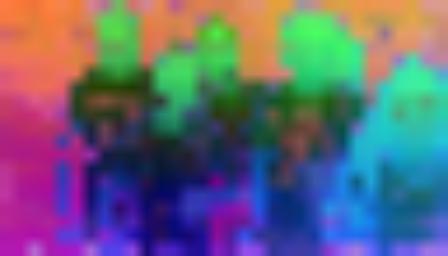} &
        \includegraphics[width=0.155\textwidth]{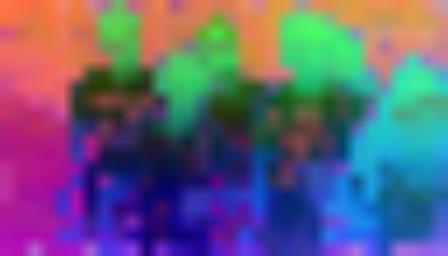} &
        \includegraphics[width=0.155\textwidth]{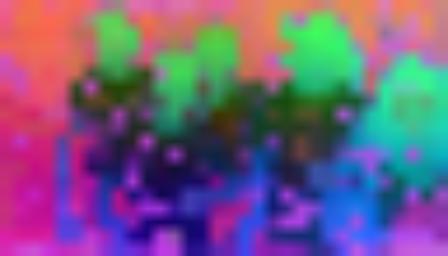} &
        \includegraphics[width=0.155\textwidth]{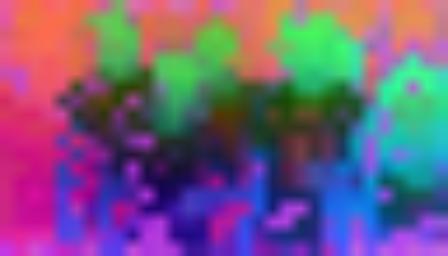} &
        \includegraphics[width=0.155\textwidth]{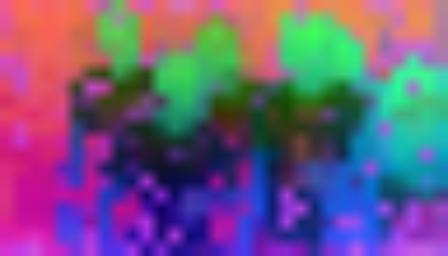} \\[3pt] 
        
        \textbf{T-1} & \textbf{T} & \textbf{T+1} & \textbf{T-1} & \textbf{T} & \textbf{T+1} \\
    \end{tabular}
    
    \caption{Visual comparison demonstrating the robustness of DINOv2 features to rain. The figure contrasts the DINOv2 feature maps extracted from (a) three consecutive frames of ground truth with those from (b) the rainy counterpart. Despite the significant visual corruption in the rainy frames, their corresponding feature maps remain remarkably temporally consistent and structurally similar to the features from the clean sequence. This stability highlights DINOv2's ability to provide reliable semantic context regardless of rain interference.}
    \label{fig:DINO} 
\end{figure*}

\noindent\textbf{Real-World Comparisons.}
Figure~\ref{fig:Real-World} presents three representative comparisons under real-world rain. In the top row, several compared methods retain a visible semi-transparent rain layer, whereas S3VD reduces this interference while preserving the fine structural details of the bricks and window. In the middle row, RainMamba produces visible degradation along the white horizontal bar, while S3VD preserves its continuity more clearly. This observation is consistent with the role of MSSF semantic guidance in maintaining salient scene structures. In the bottom row, the compared outputs retain more visible rain streaks, whereas S3VD produces fewer residual streaks and clearer background regions in the displayed example. These qualitative observations are consistent with the complementary roles of semantic guidance and spatio-temporal modeling in S3VD. We also conduct a quantitative temporal-consistency evaluation using RAFT-based Warp Error (WE). As shown in Table~\ref{tab:real_world_we}, S3VD obtains the lowest WE of 0.013, corresponding to a 55.2\% reduction from RainMamba's 0.029 and further supporting its improved temporal consistency.

\begin{table}[htbp]
\centering
\caption{Temporal-consistency comparison on real-world rainy videos.}
\label{tab:real_world_we}
\setlength{\tabcolsep}{5pt}
\resizebox{\columnwidth}{!}{%
\begin{tabular}{l|ccccc}
\toprule
Method & MSCSC & RMFD & S2VD & RainMamba & S3VD \\
\midrule
WE $\downarrow$ & 0.054 & 0.043 & 0.033 & 0.029 & \textbf{0.013} \\
\bottomrule
\end{tabular}}

\end{table}

\noindent\textbf{Analysis of DINOv2's Robustness.}
To intuitively demonstrate the robustness of the semantic priors used in our MSSF module, we visualize and compare the feature maps extracted by the pre-trained DINOv2 model from both clean and rainy video sequences. As shown in Fig.~\ref{fig:DINO}, the left half displays three consecutive clean GT frames and their corresponding DINOv2 features, while the right half shows the corrupted rainy frames and their features.

The key observation is that the feature maps extracted from the rainy frames are remarkably similar to those extracted from their clean counterparts. 
Despite the significant visual interference from rain, the DINOv2 features remain stable across different frames, consistently segmenting main objects like the crowd of people. 
This consistency, both spatially and temporally, confirms that DINOv2 can provide a reliable structural blueprint of the scene. This robust spatio-temporal cue is crucial for our S3VD, as it provides a stable target for restoration across frames, which helps to reduce temporal artifacts and enables high-fidelity results.

We further quantify this invariance on 200 paired clean--rainy frames. For the $k$-th pair, cosine similarity is averaged over $N_k$ spatially aligned feature tokens:
{
\begin{equation}
s_k=\frac{1}{N_k}\sum_{i=1}^{N_k}
\frac{\mathbf{f}^{\mathrm{rain}}_{k,i}\cdot\mathbf{f}^{\mathrm{clean}}_{k,i}}
{\lVert\mathbf{f}^{\mathrm{rain}}_{k,i}\rVert_2\,\lVert\mathbf{f}^{\mathrm{clean}}_{k,i}\rVert_2}.
\label{eq:feature_similarity}
\end{equation}
}
As reported in Table~\ref{tab:feature_similarity}, DINOv2 yields substantially higher and more stable clean--rainy similarity than ConvNeXt, quantitatively supporting its robustness to rain degradation.

\begin{table}[!h]
\centering
\caption{Clean--rainy feature similarity on the VRDS dataset~\cite{wu2023mask}.}
\label{tab:feature_similarity}
\setlength{\tabcolsep}{4pt}
\begin{tabularx}{\columnwidth}{@{}>{\raggedright\arraybackslash}p{0.50\columnwidth}|>{\centering\arraybackslash}X@{}}
\toprule
\multicolumn{1}{c|}{Feature representation} & Mean cosine similarity $\uparrow$ \\
\midrule
ConvNeXt features & $0.71\pm0.08$ \\
DINOv2 semantic features & \textbf{$0.89\pm0.04$} \\
\bottomrule
\end{tabularx}

\end{table}

\noindent\textbf{Complementarity of Semantic and Pixel Features.}
We further compare semantic-only, pixel-only, and joint-fusion configurations. As shown in Table~\ref{tab:semantic_complementarity}, using DINOv2 features alone yields 30.74~dB PSNR, substantially below the 32.35~dB of ConvNeXt pixel features. Combining pixel features with MSSF semantic guidance achieves the best result of 32.43~dB PSNR and 0.0610 LPIPS, confirming that semantic priors provide complementary structural guidance rather than replacing low-level reconstruction cues.

\begin{table}[htbp]
\centering
\caption{Complementarity between pixel and semantic representations on the VRDS dataset~\cite{wu2023mask}.}
\label{tab:semantic_complementarity}
\setlength{\tabcolsep}{3pt}
\resizebox{\columnwidth}{!}{%
\begin{tabular}{l|ccc}
\toprule
Feature configuration & PSNR $\uparrow$ & SSIM $\uparrow$ & LPIPS $\downarrow$ \\
\midrule
Semantic features only & 30.74 & 0.9291 & 0.0788 \\
Pixel features only & 32.35 & 0.9413 & 0.0642 \\
Pixel features $+$ Semantic features & \textbf{32.43} & \textbf{0.9427} & \textbf{0.0610} \\
\bottomrule
\end{tabular}}

\end{table}

\noindent\textbf{Model Complexity and Efficiency Comparison.}
We conduct a detailed analysis of the model complexity and inference efficiency of our proposed S3VD against several state-of-the-art methods. The comparison is performed on the VRDS dataset with an input resolution of $256\times256$ on a single NVIDIA RTX 4090 GPU.
For consistent accounting, S3VD has 42.08M trainable parameters and 63.08M total parameters including the frozen DINOv2-small branch, while GFLOPs and runtime are measured end to end with DINOv2 feature extraction under FP16 inference.
As Table~\ref{complexity} illustrates, S3VD achieves the fastest inference speed at 0.0089s per frame, outperforming the strong Mamba-based baseline, RainMamba (0.0112s), and significantly surpassing Transformer-based models like DRSformer (0.0381s). In terms of computational cost, S3VD maintains a low footprint with 143.58 GFLOPs, which is highly competitive and second only to RainMamba (118.99 GFLOPs) while being substantially more efficient than BasicVSR++ (1616.44 GFLOPs). While our model has a slightly larger parameter count (42.08M), it remains comparable to other high-performance models. Overall, these results validate that S3VD strikes an excellent balance, delivering state-of-the-art deraining performance without sacrificing computational efficiency, making it a practical and powerful solution.

\begin{table}[!h]
\centering
\caption{Model Complexity and Efficiency Comparisons.}
\label{complexity}
\scriptsize
\setlength{\tabcolsep}{3pt}
\resizebox{\columnwidth}{!}{
\begin{tabular}{lcccccc}
\toprule
Method   & BasicVSR++ & ESTINet & DRSformer & RainMamba & VDMamba & \textbf{Ours}\\
\midrule
Parameters(M)$\downarrow$ & \textbf{6.22} & 22.96 & 33.63 & 34.75 & \underline{12.70} & 42.08\\
FLOPs(G)$\downarrow$ & 1616.44 & 681.83 & 1101.89 & \textbf{118.99} & 389.79 & \underline{143.58}\\
Time(s/frame)$\downarrow$ & 0.051 & 0.0341 & 0.0381 & \underline{0.0112} & 0.0320 & \textbf{0.0089}\\
\bottomrule
\end{tabular}}

\end{table}

\subsection{Ablation Study}
\noindent \textbf{Ablation on Key Components.}
To validate the necessity of our proposed two key components, we conducted comprehensive ablation studies using RainMamba~\cite{wu2024rainmamba} as a baseline. We systematically examined each component's contribution through the following configurations: ``D1'' represents the baseline RainMamba with its original Coarse-to-Fine Mamba Module. ``D2'' incorporates our Multi-Scale Semantic Fusion (MSSF) module into D1. ``D3'' replaces the Coarse-to-Fine Mamba Module with our proposed Spatio-Temporal Scanning Fusion (STSF) module. Finally, our complete S3VD network integrates both innovations.

Table~\ref{ab1} presents the quantitative results in the VRDS dataset. By incorporating the MSSF module, D2 significantly enhances performance with improvements of +0.06 dB in PSNR, +0.0037 in SSIM, and -0.0049 in LPIPS compared to baseline D1. This demonstrates how MSSF's sophisticated fusion mechanism effectively leverages the complementary strengths of DINOv2's rich semantic representations and ConvNeXt's detailed pixel-level features, substantially improving the model's ability to distinguish and process complex rain patterns against varying backgrounds.
More notably, D3 delivers even more substantial gains with +0.31 dB in PSNR, +0.0047 in SSIM, and -0.0042 in LPIPS over the baseline D1. This highlights STSF's superior capability in modeling bidirectional spatio-temporal dependencies, enabling more effective correlation learning across frames and significantly enhancing temporal consistency in rain removal.
When combining both innovations in our complete S3VD model, we achieve remarkable improvements of +0.39 dB in PSNR, +0.0061 in SSIM, and -0.0074 in LPIPS compared to baseline D1. This synergistic performance boost confirms the strong complementary relationship between our MSSF and STSF, where MSSF's robust semantic understanding provides critical context that enhances STSF's spatio-temporal modeling capability. 

\begin{table}[htbp]
\centering
\caption{Ablation of key network components on the VRDS dataset~\cite{wu2023mask}.}
\label{ab1}
\setlength{\tabcolsep}{4pt}
\resizebox{\columnwidth}{!}{%
\begin{tabular}{c|cc|ccc}
\toprule
Model & MSSF & STSF & PSNR$\uparrow$ & SSIM$\uparrow$ & LPIPS$\downarrow$   \\
\midrule
D1 &  &  & 32.04 & 0.9366 & 0.0684\\
D2 & \checkmark &  & 32.10 & 0.9403 & 0.0635 \\
D3 &  & \checkmark & 32.35 & 0.9413 & 0.0642 \\
\midrule
S3VD (Ours) & \checkmark & \checkmark &\textbf{32.43} &\textbf{0.9427} &\textbf{0.0610}\\
\bottomrule
\end{tabular}}

\end{table}

\noindent\textbf{Ablation on Training Objectives.}
To examine the contributions of different training objectives and the sensitivity to their weights, we retain the pixel reconstruction loss with unit weight in all settings and use $\lambda_1$ and $\lambda_2$ to control the perceptual and DCL objectives, respectively. A zero coefficient disables the corresponding auxiliary objective. As shown in Table~\ref{tab:loss_ablation}, adding the perceptual loss to the pixel-only setting improves PSNR from 31.96 to 32.25~dB and reduces LPIPS from 0.0669 to 0.0629, while adding DCL increases PSNR to 32.18~dB and reduces LPIPS to 0.0638. Combining both objectives with $(\lambda_1,\lambda_2)=(0.3,0.1)$ achieves the best performance of 32.43~dB PSNR, 0.9427 SSIM, and 0.0610 LPIPS. These results show that the perceptual and DCL objectives provide complementary supervision. The former promotes feature-level fidelity, whereas the latter exploits temporal self-similarity to suppress temporally inconsistent degradation. The relatively small variations among the remaining weight settings further indicate that the model is not highly sensitive to modest coefficient changes around the selected values.

\begin{table}[htbp]
\centering
\caption{Loss-function ablation and weight sensitivity on VRDS~\cite{wu2023mask}.}
\label{tab:loss_ablation}
\renewcommand{\arraystretch}{1.08}
\setlength{\tabcolsep}{2pt}
\begin{tabular*}{\columnwidth}{@{\extracolsep{\fill}}cc|ccc@{}}
\toprule
$\lambda_1$ & $\lambda_2$ & PSNR $\uparrow$ & SSIM $\uparrow$ & LPIPS $\downarrow$ \\
\midrule
0 & 0 & 31.96 & 0.9388 & 0.0669 \\
0.3 & 0 & 32.25 & 0.9416 & 0.0629 \\
0 & 0.1 & 32.18 & 0.9409 & 0.0638 \\
\textbf{0.3} & \textbf{0.1} & \textbf{32.43} & \textbf{0.9427} & \textbf{0.0610} \\
\midrule
0.1 & 0.1 & 32.34 & 0.9420 & 0.0621 \\
0.3 & 0.05 & 32.38 & 0.9423 & 0.0616 \\
0.3 & 0.2 & 32.37 & 0.9425 & 0.0613 \\
0.5 & 0.1 & 32.35 & 0.9424 & 0.0612 \\
\bottomrule
\end{tabular*}

\end{table}

\begin{table}[h!]
\renewcommand\arraystretch{1}
\centering
\caption{Ablation of semantic extractors on the VRDS dataset~\cite{wu2023mask}.}
\label{ab2}
\resizebox{0.48\textwidth}{!}{
\begin{tabular}{l|cccccc}
\toprule
\multicolumn{1}{c|}{Model} & PSNR$\uparrow$ & SSIM$\uparrow$ & LPIPS$\downarrow$ & Param$\downarrow$ & FLOPS$\downarrow$ & Runtime$\downarrow$\\
\midrule
DINOv2-small (Ours) & 32.43 & 0.9427 & 0.0610 & 63.08 & 143.58 & 0.0089\\
DINOv2-small with register & 32.39 & 0.9417 & 0.0622 & 63.08  & 143.58 & 0.0078\\
DINOv2-base & 32.53 & 0.9431 & 0.0618 & 127.63 & 248.42 & 0.0094\\
DINOv2-large & 32.58 & 0.9422 & 0.0618 & 345.40 & 602.17 & 0.0143\\
DINOv1-small & 32.34 & 0.9410 & 0.0632 & 63.30 & 129.43 & 0.0076 \\
CLIP-visual & 32.42 & 0.9413 & 0.0627 & 127.76 & 193.21 & 0.0082 \\
CLIP-text & 32.26 & 0.9391 & 0.0662 & 38.61 & 106.45 & 0.0065 \\
\bottomrule
\end{tabular}%
}

\end{table} 

\noindent\textbf{Ablation on Semantic Extractors.}
To validate our choice of DINOv2 as the semantic prior and to investigate the impact of different semantic extractors on deraining performance, we conducted a comprehensive ablation study. We integrated several powerful pre-trained vision models into our S3VD framework as the semantic guidance module, including various sizes of DINOv2 (DINOv2-small, DINOv2-base, DINOv2-large) and the widely used text-image alignment model, CLIP (CLIP-visual, CLIP-text).
The quantitative results, presented in Table~\ref{ab2}, clearly indicate that DINOv2-based variants consistently outperform their CLIP-based counterparts in deraining accuracy (PSNR). We attribute this performance gap to the inherent nature of these models. CLIP's training objective, which focuses on aligning images with text, can introduce a bias that is detrimental to pixel-level restoration tasks where precise spatial detail is paramount. In contrast, DINOv2, trained via self-supervision on curated image data, demonstrates superior robustness to visual degradation and a stronger capacity for capturing the global structure and physical semantics of scenes. Among the DINOv2 variants, DINOv2-small strikes an optimal balance between performance, efficiency (FLOPs, Parameters), and its resilience to noise and occlusions. Therefore, we selected DINOv2-small as the default semantic extractor for our S3VD framework.

\begin{table}[htbp]
\centering
\caption{Results of different setting of query in the MSSF on the VRDS dataset~\cite{wu2023mask}.}
\label{ab3}
\setlength{\tabcolsep}{4pt}
\resizebox{\columnwidth}{!}{%
\begin{tabular}{c|ccc}
\toprule
Method & PSNR$\uparrow$ & SSIM$\uparrow$ & LPIPS$\downarrow$   \\
\midrule
single frame &32.22 &0.9420 &0.0622\\
cross-frame average &32.39 &0.9425 &0.0611 \\
cross-frame concatenation (Ours) &\textbf{32.43} &\textbf{0.9427} &\textbf{0.0610}\\
\bottomrule
\end{tabular}}

\end{table}

\noindent\textbf{Ablation on Multi-Scale Semantic Fusion Module.}
To evaluate the fusion mechanism between DINOv2 semantic priors and temporal information in our MSSF module, we conducted ablation experiments comparing three query formation strategies, as shown in Table~\ref{ab3}. Our cross-frame concatenation approach achieved the best performance (32.43 dB PSNR, 0.9427 SSIM, 0.0610 LPIPS), outperforming single-frame queries by 0.21 dB in PSNR, 0.0007 in SSIM, and 0.0012 in LPIPS. The performance gap reveals important insights about temporal information utilization: single-frame queries process frames independently without leveraging inter-frame relationships, while cross-frame averaging incorporates temporal information but tends to dilute distinctive frame features. In contrast, our cross-frame concatenation strategy preserves each frame's unique characteristics while establishing effective temporal correlations, enabling the model to comprehensively capture both spatial variations within frames and the temporal evolution of rain patterns across consecutive frames--a critical capability for achieving high-quality rain removal with strong temporal consistency.

To further assess the necessity of hierarchical multi-scale fusion, we progressively increase the fused ConvNeXt levels from the deepest feature $L_4$ to the complete set $L_1$--$L_4$, while keeping all other settings unchanged. As shown in Table~\ref{tab:mssf_scales}, the single-scale $L_4$ configuration achieves 32.19~dB PSNR, 0.9408 SSIM, and 0.0641 LPIPS. Incorporating $L_3$ and then $L_2$ steadily improves PSNR to 32.28 and 32.36~dB, respectively, while reducing LPIPS to 0.0628 and 0.0617. The full four-scale configuration performs best, reaching 32.43~dB PSNR, 0.9427 SSIM, and 0.0610 LPIPS, corresponding to a 0.24~dB PSNR gain and a 0.0031 LPIPS reduction over the single-scale setting. These consistent improvements show that the hierarchy levels provide complementary rather than redundant information. Deeper features supply coarse semantic context and larger receptive fields, whereas higher-resolution features preserve edges and fine spatial details. Their joint use therefore enables MSSF to combine structure-aware guidance with detail-sensitive restoration.

\begin{table}[htbp]
\centering
\caption{Ablation of fused feature levels in MSSF on the VRDS dataset~\cite{wu2023mask}.}
\label{tab:mssf_scales}
\setlength{\tabcolsep}{4pt}
\begin{tabular*}{\columnwidth}{@{\extracolsep{\fill}}lccc@{}}
\toprule
Fused levels & PSNR$\uparrow$ & SSIM$\uparrow$ & LPIPS$\downarrow$ \\
\midrule
$L_4$ & 32.19 & 0.9408 & 0.0641 \\
$L_3+L_4$ & 32.28 & 0.9415 & 0.0628 \\
$L_2+L_3+L_4$ & 32.36 & 0.9422 & 0.0617 \\
$L_1+L_2+L_3+L_4$ & \textbf{32.43} & \textbf{0.9427} & \textbf{0.0610} \\
\bottomrule
\end{tabular*}

\end{table}

\begin{table}[htbp]
\centering
\caption{Ablation off semantic--pixel fusion modules on the VRDS dataset~\cite{wu2023mask}.}
\label{tab:fusion_operator}
\setlength{\tabcolsep}{1.5pt}
\resizebox{\columnwidth}{!}{%
\begin{tabular}{l|cc|ccc}
\toprule
Fusion operator & Params (M) & GFLOPs & PSNR$\uparrow$ & SSIM$\uparrow$ & LPIPS$\downarrow$ \\
\midrule
Concatenation $+$ $1\times1$ Conv & 40.31 & 132.75 & 32.17 & 0.9406 & 0.0638 \\
Bidirectional Mamba fusion & 42.36 & 146.91 & 32.31 & 0.9418 & 0.0622 \\
Cross-attention (MSSF) & 42.08 & 143.58 & \textbf{32.43} & \textbf{0.9427} & \textbf{0.0610} \\
\bottomrule
\end{tabular}}

\end{table}

To compare different semantic--pixel fusion operators, we evaluate channel concatenation, bidirectional Mamba fusion, and cross-attention under the same settings. As shown in Table~\ref{tab:fusion_operator}, cross-attention achieves the best performance with 32.43~dB PSNR, 0.9427 SSIM, and 0.0610 LPIPS, while using slightly fewer parameters and GFLOPs than bidirectional Mamba fusion.

\noindent\textbf{Ablation on Spatio-Temporal Scanning Fusion Module.}
In the STSF module, we conducted ablation studies to evaluate how different arrangements of spatial and temporal Mamba blocks affect rain removal performance, as shown in Table~\ref{ab4}. Our S+T configuration (Spatial followed by Temporal) achieved the best results with 32.43 dB PSNR, 0.9427 SSIM, and 0.0610 LPIPS, significantly outperforming all alternatives. This arrangement excels in video deraining because it first captures the spatial characteristics of rain streaks and raindrops within individual frames before modeling their temporal dynamics across consecutive frames--a critical sequence for distinguishing between static background elements and transient rain patterns.
The T+T combination ranked second with 32.31 dB PSNR, 0.9414 SSIM, and 0.0634 LPIPS, confirming the importance of temporal modeling in tracking rain streak trajectories and raindrop movements across frames. However, without initial spatial feature extraction, this configuration struggles to differentiate rain elements from complex background textures, resulting in less effective rain removal. The S+S configuration placed third (32.13 dB PSNR, 0.9407 SSIM, 0.0625 LPIPS), effectively identifying spatial rain patterns but failing to leverage temporal coherence essential for removing dynamic rain elements that move across frames.
Notably, the T+S arrangement performed worst (32.05 dB PSNR, 0.9402 SSIM, 0.0638 LPIPS), with a significant gap of 0.38 dB PSNR compared to our S+T design. This demonstrates that attempting to model temporal rain dynamics before establishing strong spatial representations disrupts the network's ability to effectively identify and remove rain patterns. These findings validate our STSF design, confirming that first establishing spatial rain characteristics before analyzing their temporal evolution creates an optimal framework for comprehensive rain removal in video sequences, effectively handling both stationary raindrops on surfaces and dynamic rain streaks in motion.

\begin{table}[htbp]
\centering
\caption{Results of different combinations of spatio and temporal Mamba blocks on the VRDS dataset~\cite{wu2023mask}.}
\label{ab4}
\centering
\begin{tabularx}{\columnwidth}{@{}>{\centering\arraybackslash}p{0.24\columnwidth}|*{3}{>{\centering\arraybackslash}X}@{}}
\toprule
Method & PSNR$\uparrow$ & SSIM$\uparrow$ & LPIPS$\downarrow$ \\
\midrule
T+S & 32.05 & 0.9402 & 0.0638 \\
S+S & 32.13 & 0.9407 & 0.0625 \\
T+T & 32.31 & 0.9414 & 0.0634 \\
S+T (Ours) & \textbf{32.43} & \textbf{0.9427} & \textbf{0.0610} \\
\bottomrule
\end{tabularx}

\end{table}

\noindent\textbf{Ablation on STSF Depth.}
To determine an appropriate stage-wise depth, we evaluate four STSF configurations while keeping the remaining architecture and training settings unchanged. As shown in Table~\ref{tab:stsf_depth}, increasing the depth from $(1,2,1)$ to $(2,2,2)$ and then to $(2,3,2)$ consistently improves all restoration metrics. Compared with $(2,2,2)$, the adopted $(2,3,2)$ configuration improves PSNR by 0.09~dB and SSIM by 0.0007 while reducing LPIPS by 0.0011, with only 6.0\% more GFLOPs and 8.5\% more runtime. By contrast, further deepening the network to $(3,4,3)$ increases GFLOPs and runtime by 18.3\% and 19.1\%, respectively, but provides only a 0.02~dB PSNR gain. This saturation indicates that $(2,3,2)$ offers a favorable accuracy--complexity trade-off. The additional middle-stage module operates on lower-resolution features with a larger effective receptive field, strengthening spatio-temporal dependency aggregation without repeatedly increasing high-resolution processing.

\begin{table}[htbp]
\centering
\caption{Ablation of stage-wise STSF depth on the VRDS dataset~\cite{wu2023mask}.}
\label{tab:stsf_depth}
\setlength{\tabcolsep}{1pt}
\begin{tabular*}{\columnwidth}{@{\extracolsep{\fill}}l|cccccc@{}}
\toprule
Depth & Params$\downarrow$ & GFLOPs$\downarrow$ & Time$\downarrow$ & PSNR$\uparrow$ & SSIM$\uparrow$ & LPIPS$\downarrow$ \\
\midrule
$(1,2,1)$ & \textbf{35.42} & \textbf{118.62} & \textbf{0.0074} & 32.18 & 0.9405 & 0.0642 \\
$(2,2,2)$ & 39.76 & 135.47 & 0.0082 & 32.34 & 0.9420 & 0.0621 \\
$\underline{(2,3,2)}$ & \underline{42.08} & \underline{143.58} & \underline{0.0089} & \underline{32.43} & \underline{0.9427} & \underline{0.0610} \\
$(3,4,3)$ & 48.91 & 169.84 & 0.0106 & \textbf{32.45} & \textbf{0.9428} & \textbf{0.0609} \\
\bottomrule
\end{tabular*}

\end{table}

\begin{table}[htbp]
\centering
\caption{Component-wise ablation of DG-Mamba on the VRDS dataset.}
\label{ab5}
\centering
\setlength{\tabcolsep}{2.5pt}
\begin{tabular*}{\columnwidth}{@{\extracolsep{\fill}}cc|ccc@{}}
\toprule
Shared core & Direction gates & PSNR$\uparrow$ & SSIM$\uparrow$ & LPIPS$\downarrow$ \\
\midrule
-- & -- & 32.12 & 0.9410 & 0.0624 \\
\checkmark & -- & 32.24 & 0.9417 & 0.0619 \\
-- & \checkmark & 32.29 & 0.9421 & 0.0616 \\
\checkmark & \checkmark & \textbf{32.43} & \textbf{0.9427} & \textbf{0.0610} \\
\bottomrule
\end{tabular*}

\end{table}

\noindent\textbf{Ablation on DG-Mamba.}
To disentangle the contributions of its two key designs, we conduct a $2\times2$ component-wise ablation of the shared Conv1d/SSM core and direction-specific gates while keeping all other settings unchanged. As shown in Table~\ref{ab5}, parameter sharing alone improves the Bi-Mamba baseline from 32.12 to 32.24~dB PSNR and reduces LPIPS from 0.0624 to 0.0619, whereas direction-specific gating alone increases PSNR to 32.29~dB and reduces LPIPS to 0.0616. Combining both designs achieves the best performance of 32.43~dB PSNR, 0.9427 SSIM, and 0.0610 LPIPS. These results demonstrate the complementary roles of the two designs. The shared core encourages the two scanning directions to learn common transition dynamics and avoids redundant modeling, while the independent gates retain content-adaptive control over forward and backward contextual information.

\subsection{Limitations}
Although S3VD performs favorably on average, its restoration quality decreases under very dense rain, severe motion blur, large raindrop occlusion, and rapid camera motion. Dense rain and large occlusions reduce the available background evidence, severe blur weakens fine spatial features, and rapid motion disrupts the fixed-coordinate temporal correspondence used by TMB. These conditions may result in residual rain, over-smoothed details, or local temporal inconsistencies, motivating future work on motion-aware alignment and occlusion-aware temporal aggregation.

\section{Conclusion}
\label{sec:conclusion}
In this paper, we propose S3VD, a novel video deraining framework that effectively addresses the limitations of existing methods in semantic perception and spatio-temporal consistency. We introduce a Multi-Scale Semantic Fusion (MSSF) Module that integrates semantic priors from DINOv2 to guide precise feature representation, enhancing robustness against severe degradation. Furthermore, we design a Spatio-Temporal Scanning Fusion (STSF) Module to effectively capture the spatio-temporal correlation within rainy videos. STSF consists of Spatial Mamba Blocks (SMB) and Temporal Mamba Blocks (TMB) arranged sequentially, which simulate spatial and temporal relationships in video sequences separately. Each Mamba block incorporates our novel Decoupled-Gating Mamba (DG-Mamba) layer. DG-Mamba employs dual gating branches to adaptively regulate forward/backward scanning, while sharing 1D convolution and SSM parameters, enabling complementary spatio-temporal feature fusion. Extensive experimental evaluations demonstrate that our S3VD consistently outperforms current state-of-the-art approaches on multiple benchmark datasets for video deraining, effectively removing raindrops and rain streaks while preserving spatio-temporal coherence and structural details. Our work makes significant contributions to the field of video deraining by pioneering architectural innovations to the Mamba framework, adapting its 1D sequence processing capabilities to better suit the unique challenges of 3D video tasks, and establishing a new paradigm for efficient, semantic-aware spatio-temporal modeling in video restoration.

\section*{Acknowledgments}
This research was financially supported by the Fundamental and Interdisciplinary Disciplines Breakthrough Plan of the Ministry of Education of China (JYB2025XDXM906), National Natural Science Foundation of China (62501189), Natural Science Foundation of Heilongjiang Province of China for Excellent Youth Project (YQ2024F006), and the Hubei Provincial Key Research and Development Program under Grant (2024BAB039).

\bibliographystyle{wileyNJD-AMA}
\bibliography{main}

\end{document}